# VIS-CRF, A CLASSICAL RECEPTIVE FIELD MODEL FOR VISION

Nasim Nematzadeh, David M.W. Powers, Trent Lewis

## 1. Introduction

Over the last decade, a variety of new neurophysiological experiments have led to new insights as to how, when and where retinal processing takes place, and the nature of the retinal representation encoding sent to the cortex for further processing. Based on these neurobiological discoveries, in our previous work, we provided computer simulation evidence to suggest that Geometrical illusions are explained in part, by the interaction of multiscale visual processing performed in the retina. The output of our retinal stage model, named Vis-CRF[1], is presented here for a sample of natural image and for several types of Tilt Illusion, in which the final tilt percept arises from multiple scale processing of Difference of Gaussians (DoG) and the perceptual interaction of foreground and background elements (Nematzadeh and Powers, 2019; Nematzadeh, 2018; Nematzadeh, Powers and Lewis, 2017; Nematzadeh, Lewis and Powers, 2015).

The bioplausible model used is similar to Robson's (1983) which found the retinal ganglion cell (GC) response to patterns (images). He demonstrated that a model of a ganglion cell receptive field (RF) with antagonistic center and surround organization of DoG can predict the cell response to any arbitrary stimulus pattern. What is needed is to find the convolution between the pattern's luminance function with the differences of Gaussians as the *weighting function of ganglion cells RFs*. One of the first models for foveal retinal vision was proposed by Lindeberg and Florack (1994) and our model is largely inspired by it. Their model is based on simultaneous sampling of the image at all scales, and the edge map at multiple scales in our model is generated in a similar way. We show that the weighting function of the ganglion cell (GCs) has been properly set in our Vis-CRF model which resulted in an accurate *VISION* model simulating the *lateral inhibition* of the simple cells. Our model is *unique* in the sense that for the first time we have demonstrated a mathematical model that actually predicts the precise distortion angles a human will see in some optical illusions, and thus empirically testable.

This study reflects the core of the lead author's Ph.D. research (Nematzadeh, 2018) which focused on illusions in which lines of tiles or squares were perceived to bend or tilt (such as Café Wall and checkerboard illusions). This research is interdisciplinary across human and computer vision, with low-level models grounded in neuroscience and high-level models grounded in psychology. Our work is not about computational intelligence (making computer more intelligent) but human intelligence (explaining human's cognition), so this work is out of the domain of artificial/computational intelligence.

---

[1] The Classical Receptive Field (CRF) model for Vision (Vis)

In this text, we start with an introduction to multiscale representation in vision and in computer vision research (Section 2). Then, we move to explain the Differences of Gaussians as a bioplausible filtering approach for modelling lateral inhibition of simple cells in the retina and in the cortex (Section 3). Succeeding this, a formal description of Vis-CRF model is provided followed by various parameters involved in the DoG-based model. Next, an edge map representation of the DoG response to a stimulus (a cropped section of a Café Wall pattern) will be described with a deeper investigation of the effect of the parameters on this response (Section 4). We continue by presenting a flowchart of the model and an introduction to an image-processing pipeline designed for tilt analysis by further exploration of the edge map for the investigated Café Wall illusion (Section 5) and for the "specific application of Geometrical illusions" in Section 6 (A complete investigation of tilt analysis can be found in (Nematzadeh, 2018)). In Section 6, we utilise our Vis-CRF model on a wide range of Tile/Tilt illusions including a few samples generated by the lead author. We conclude this work in Section 7 by the further extension to our model, highlighting the similarity of the edge map representation to Marr's primal sketch and his theory of vision shown for a scenery stimulus/image.

## 2. Multiscale representation in vision and vision models

A model of multiscale transforms for the processing and projections in the visual cortex of mammals as well as the visual processing inside the retina was suggested by psychophysical and physiological findings (Field and Chichilnisky, 2007; Gollisch and Meister, 2010; Gauthier, Field, Sher et al., 2009). Kuffler was a pioneer in recording the activity of retinal ganglion cells (RGCs) that make up the optic nerve and identified two types of center-surround cells (1952). Hubel and Wiesel also devised and performed many pioneering experiments that increased our understanding of cortical visual processing (Hubel and Wiesel, 1962). Moreover, Daugman developed a mathematical model of cortical cells using Gaussian windows modulated by a sinusoidal wave for this impulse response that predicted the specific spatial orientation tunings of these cells from the dilation of modulated Gabor functions (Daugman, 1980).

The need to extract multiscale image information in modelling the visual mechanism in Computer Vision (CV) applications has been established by many researchers in the field (Rosenfeld and Thurston, 1971; Marr and Hildreth, 1980; Marr, 1982; Burt and Adelson, 1983). Some of these ideas have later been subsumed by the wavelet paradigm. In a multiresolution algorithm by Burt and Adelson (1983), the search typically started from coarse to fine-scale, processing low-resolution images first and then zooming selectively into the fine scales of visual data. Mallat (1996) highlighted the impacts of wavelets for low-level vision in multiresolution searches, multiscale edge detection, and texture discrimination.

Pyramidal image representations and scale-invariant transforms (Lowe, 1999) are well-matched to human visual encoding and do not need image partitioning like JPEG-DCT (Taubman and Marcellin, 2012). A scale-space analysis is an emergent result of image decomposition as differences between pairs of scaled filters with different parameterizations, subsuming the Laplacian and Difference of Gaussian filters (LoG/DoG) (Jacques, Duval, Chaux et al., 2011; Lindeberg, 2011).

Note further that, self-organizing models that demonstrate patterns of edge detectors at particular angles are well established (von der Malsburg, 1973). Higher-level spatial aggregation of regularly spaced spots or edges in turn automatically gives rise to analogues of DCT and DWT type bases, the latter with localization determined by the higher level lateral interaction functions or the constraints of an underlying probabilistic connectivity model (Powers, 1983).

In our visual system, light is focused onto receptors that transduce it to neural impulses which further processed in the retina (Smith, 2003). Then the middle layer of the retina, which is the focus of our study, enhances the neural signals through the process of '*lateral inhibition*' (Ratliff, Knight and Graham, 1969), causing an activated nerve cell in the middle layer of the retina to decrease the ability of its neighbouring neurons to become active. This '*biological convolution*' with its specific '*Point Spread Function*' (PSF) improves the ability of the eye to see the world but also, as we show, leads to optical illusions. The effect of center-surround processing and lateral inhibition on an indistinctly defined edge with a gradual change from dark to light is that it reinforces transition between the light and dark, making them appear more abrupt due to the appearance of overshoots and undershoots (Smith, 2003). This results in the sharpening of edges and facilitates visual tasks.

The first layer of the retina has a nonlinear mechanism for retinal gain control, flattening the illumination component, and making it possible for the eye to see under poor light condition (Smith, 2003). Thus, the *lateral inhibition* in the middle layer of the retina evidences both a *bandpass filtering* property and an *edge enhancement* capability. After the final layer, ganglion cell axons exit the eye and carry the visual signals to the cortex. In addressing Brightness/Lightness illusions as well as Geometrical/Tilt Illusions, it appears that the mechanisms of the first two layers of the retina provide very important clues in understanding illusions involving edges.

The *Contrast Sensitivity* of retinal ganglion cells can be modelled based on Classical Receptive Field (CRFs) with their circular center and surround antagonism, which use differences and second differences of Gaussian (Enroth-Cugell and Robson, 1966; Rodieck and Stone, 1965) or Laplacian of Gaussian (Ghosh, Sarkar and Bhaumik, 2007) to reveal edge information. Marr and Hildreth (1980) proposed an approximation of LoG with DoG based on a specified ratio of the standard deviations ($\sigma$) of the Gaussians (Enroth-Cugell and Robson, 1966; Rodieck and Stone, 1965). Additionally, Powers (1983) showed that DoG models can themselves result from a simple biophysical model of ontogenesis and can usefully approximate the interaction functions proposed in a variety of neural models.

Our visual perception of a scene starts by extracting the multiscale edge map, and a bioplausible implementation of contrast sensitivity of retinal RFs using DoG filtering produces a stack of multiscale outputs (Romeny, 2003). In retinal encoding, what is sent to the brain is a stack of images (impulse responses) or a scale-space, not a single image. One of the first models for foveal retinal vision was proposed by Lindeberg and Florack (1994) and our model is inspired by it. Their model is based on the simultaneous sampling of the image at all scales, and the edge map in our model is generated in a similar way. The extraction of an edge map is an essential and primitive task in most image processing applications, but according to Marr's *raw primal sketch* to *full primal sketch* and perception of a 3D view of the world (Marr and Hildreth, 1980) we need further information for our final perception. There are also possibilities of the involvement of higher-order derivatives of Gaussians, which can be seen in retinal to cortical visual processing models such as: (Marr and Hildreth, 1980; Young, 1985; Young, 1987; Lourens, 1995; Ghosh, Sarkar and Bhaumik, 2007), but there is no biological evidence for them.

Furtheremore, Powers (1983) also proposed an ontogenetic Bernoulli-like model showing that an appropriate lateral interaction function can self-organize, and can approximate many existing mathematical models, including DoG models and LoG models (emergent as two levels of DoG processing) noting that processing is not particularly sensitive to the parameterization or shape of the filter functions used. Indeed, cluster-level aggregates of Powers' Bernoulli model approximate to Poisson and Gaussian models.

The developed model (Vis-CRF) has some similarities with the *Brightness Assimilation and Contrast* theory. An early model covering both brightness contrast and assimilation was developed by Jameson (1985) based on DoG filters with multiple spatial scales. In a later paper, Jamson and Hurvich (1989) point out that parallel processing occurs as the result of the simultaneous appearance of sharp edges and mixed colour that define delimited regions. Their suggestion for the source of contrast and assimilation is that the contrast effect happens when the stimulus components are relatively large in size compared to the center of the filter, and the assimilation effect happens when components of the stimulus are small compared to the filter center.

Recent physiological findings on retinal ganglion cells (RGCs) have dramatically extended our understanding of retinal processing. Previously, it was believed that retinal lateral inhibition could not be the cause of the Café Wall illusion because the effect is highly directional and arises from the preserving of both orientation as well as spatial frequency. Neuro-computational implementations of eye models have been proposed (Young, 1987; Romeny, 2003) based on the biological findings (Shapley and Perry, 1986) that take into account for the size variation of RGCs due to changes of the eccentricity and dendritic field size. Field and Chichilnisky (2007) published a detailed study about the circuitry and coding of the information processing inside the retina, and noted the existence of at least 17 distinct retinal ganglion cell types and their specific role in visual information encoding.

Some RGCs have orientation selectivity similar to the cortical cells (Barlow and Hill, 1963; Weng, Sun and He, 2005). Additionally, there is evidence that some retinal cells including horizontals and amacrine interneurons can have an elongated surround beyond the CRF size. This leads to orientation-selective models of the eye, called retinal non-CRFs (nCRFs) (Carandini, 2004a; Cavanaugh, Bair and Movshon, 2002; Wei, Zuo and Lang, 2011).

The proposed underlying mechanisms of retinal multiscale processing, from fine to coarse scales, is strongly supported by the diversity of intra-retinal circuits, and the different types of RGCs available (Field and Chichilnisky, 2007; Gauthier, Field, Sher et al., 2009). Furthuremore, there are variations in the size of each individual RGC in relation to the retinal eccentricity and their distance from the fovea (Lourens, 1995). This feature indicates a high likelihood of the involvement of retinal/cortical simple cells and their early visual processing in revealing tilt cues inside tile illusion patterns (Café Wall in particular).

## 3. The bioplausible Difference of Gaussian model

### 3.1. Formal description and parameters

The features of our bioplausible model should have the characteristics of human early visual processing. Based on numerous physiological studies, for example (Field and Chichilnisky, 2007; Gauthier, Field, Sher et al., 2009; Gollisch and Meister, 2010), there is diversity in the receptive field types and sizes inside the retina, resulting in the multiscale encoding of the visual scene. This retinal representation is believed to be *scale-invariant* in general, and an adaptation mechanism for the receptive field sizes to some textural elements is present inside our field of view (Romeny, 2008; Craft, Schutze, Niebur et al., 2007).

Applying a Gaussian filter to an image results in its blurred version. For a 2D signal such as image *I*, the DoG output (one scale of the edge map) of our retinal Vis-CRF model with center-surround organization is given by:

$$\Gamma_{\sigma, s\sigma}(x,y) = I * 1/2\pi\sigma^2 \exp[-(x^2+y^2)/(2\sigma^2)] - I * 1/2\pi(s\sigma)^2 \exp[-(x^2+y^2)/(2s^2\sigma^2)] \qquad (1)$$

where $x$ and $y$ are the distance from the origin in the horizontal and vertical axes respectively and $\sigma$ is the standard deviation/radius of center Gaussian ($\sigma_c$) (* is the convolution operator). As shown in (2) $s\sigma$ indicates the standard deviation of the surround Gaussian ($\sigma_s = s\sigma$). $s$ is referred to as *Surround ratio* in our model.

$$s = \sigma_{surround} / \sigma_{center} = \sigma_s / \sigma_c \qquad (2)$$

As indicated in (1) and (2), increases in the value of $s$, leads to a wider area of surround suppression, although the height of the surround Gaussian declines. This is because the inhibitory surround exactly balances the excitatory center that results in the area under the curve equating to zero. Further explanation for the effect of this parameter on the shape of the DoG filter is provided in Section 3.2.

In addition to the $s$ factor, the filter size is another essential parameter to be considered for the model. The DoG is only applied within a window where the value of both Gaussians are insignificant outside the window (less than 5% for the surround Gaussian). A parameter called *Window ratio* ($h$) is defined to control window size. The window size is determined based on this parameter ($h$) and the standard deviation of the center Gaussian ($\sigma_c$), as given in (3):

$$Window\ size = h \times \sigma_c + 1 \qquad (3)$$

Parameter $h$ determines how much of each Gaussian (center and surround) is included inside the DoG filter (+1 as given in (3) guarantees a symmetric filter). Further explanation for the effect of this parameter on the shape of the DoG filter is provided in Section 3.3. For the experimental results, in order to capture both excitation and inhibition effects of the retinal cell activities, the *Window ratio* was primarily set to 8 ($h$=8) in our investigations.

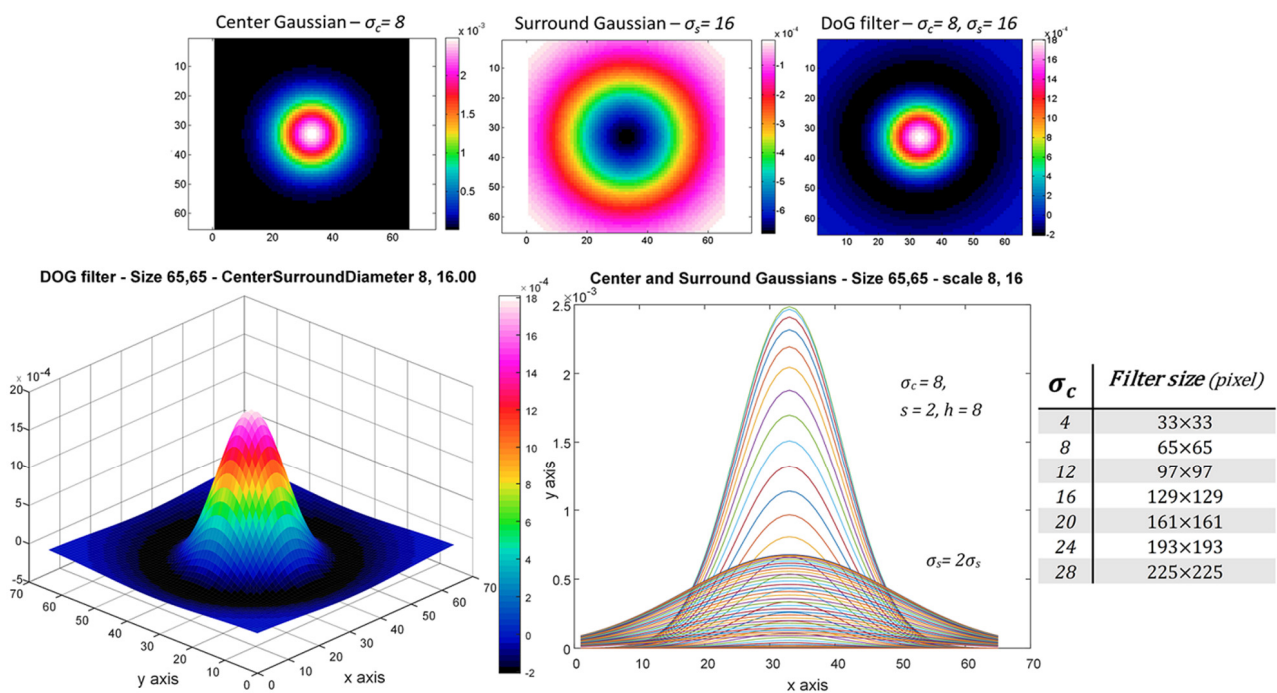

**Figure 1** *Top*: 2D representation of center and surround Gaussians with standard deviations of $\sigma_c$=8 and $\sigma_s$=16 (*Surround ratio* of $s$=2), and the result of the Difference of Gaussian (DoG) filter in a jetwhite colormap. *Bottom* (*Left*): 3D surface of the above DoG filter, and Bottom (*Right*): The relationship between *Window ratio* ($h$) and the coverage of the center and surround Gaussians in the DoG filter (Nematzadeh, 2018).

Figure 1 (Top) illustrates a 2D representation of two separate Gaussian filters for center and surround with their difference resulting in the DoG filter, using the jetwhite colormap[2] (Powers, 2016). Figure 1 (Bottom-left) shows a 3D surface of the DoG filter, and (Bottom-right) the connection of *Window ratio* ($h$) to the shape of the DoG filter ($\sigma_c$ and $\sigma_s$) in our model.

The Laplacian of Gaussian can be found by calculating the Differences between two DoGs estimating the second derivative of Gaussian. To model the receptive field of retinal Ganglion Cells, DoG filtering (Rodieck and Stone, 1965; Enroth-Cugell and Robson, 1966) is a good approximation of Laplacian of Gaussian (Logvinenko and Kane) if the ratio of dispersion of center to surround is close to 1.6 (Earle and Maskell, 1993; Marr and Hildreth, 1980) ($s \approx 1.6 \approx \varphi$, the ubiquitous *Golden Ratio*).

### 3.2. *Surround ratio*($s$) and DoG filters

The reason for using $s$=2.0 or 1.6 ($s=\sigma_s/\sigma_c$: *Surround ratio*) in our model is that the ratio of the size of center : surround Gaussians equal to 1:1.6 – 2.0 is a typical range for modelling simple cells. The ratio of 1:1.6 is given by Marr and Hildreth (1980) for modelling retinal GCs in general, and Earle and Maskell (1993) used this ratio for DoG modelling specifically to explain the Café Wall illusion. Other ratios such as 1:5.0 have been used by Lulich and Stevens (1989). We have tested a range of 1.4 to up to 8.0 for the s parameter in our implementations. However, we reiterate that our model is insensitive to changes in the center-surround ratio in the range tested from 1:1.4 to 1:8.0. Our model with its 1.6-2× ratio relates to simple cells, for example, midget bipolar cells, whereas a ratio of 4-8× relates to complex cells (Schiller, 2010; see p. 17090).

One may expect for the geometry of the DoG filters in the model to closely resemble actual ganglion cells structure and how this might affect the model's performance. For a bioplausible implementation of simple cells, we need biological evidence which is usually determined based on fitting models of either two or three Gaussians to the response of specific retinal GCs (e.g. midget and parasol) or different bipolar cells. Evidence largely comes from non-humans including cats and monkeys. As noted in the conclusion of Linsenmeier, Frishman, Jakiela et al. (1982) for the cat's retina, different values for this ratio were reported in the literature. For example, "Cleland, Levick and Sanderson (1973) found that the surround diameter was on the average 5.13 and 4.25 times larger than the center diameter in sustained and transient cells respectively; Enroth-Cugell and Robson (1966) found a value of 6.74 for their X cells; and in the LGN So and Shapley (1981) have reported values of 4.58 for X and 7.99 for a small sample of Y cells. Our value for X cells is 3.88, which is not too different, but for Y cells we found an average of only 1.49" (p.1182).

A recent human physiology study by Field and Chichilnisky (2007) indicates there are at least 17

---

[2]https://au.mathworks.com/matlabcentral/fileexchange/48419-jetwhite-colours-/content/jetwhite.m

distinct GCs in the retina. Each GCs type has a diverse range of sizes in relation to the eccentricity of the neurons and the distance from the fovea. Even the ON and OFF cells of each GCs type have a different size. Also, the possibility of simultaneous activations of a group of GCs (combined activity) in the retina by the output of amacrine cells is noted in the literature (Barlow, Derrington, Harris et al., 1977; Frishman and Linsenmeier, 1982; Roska and Werblin, 2003).

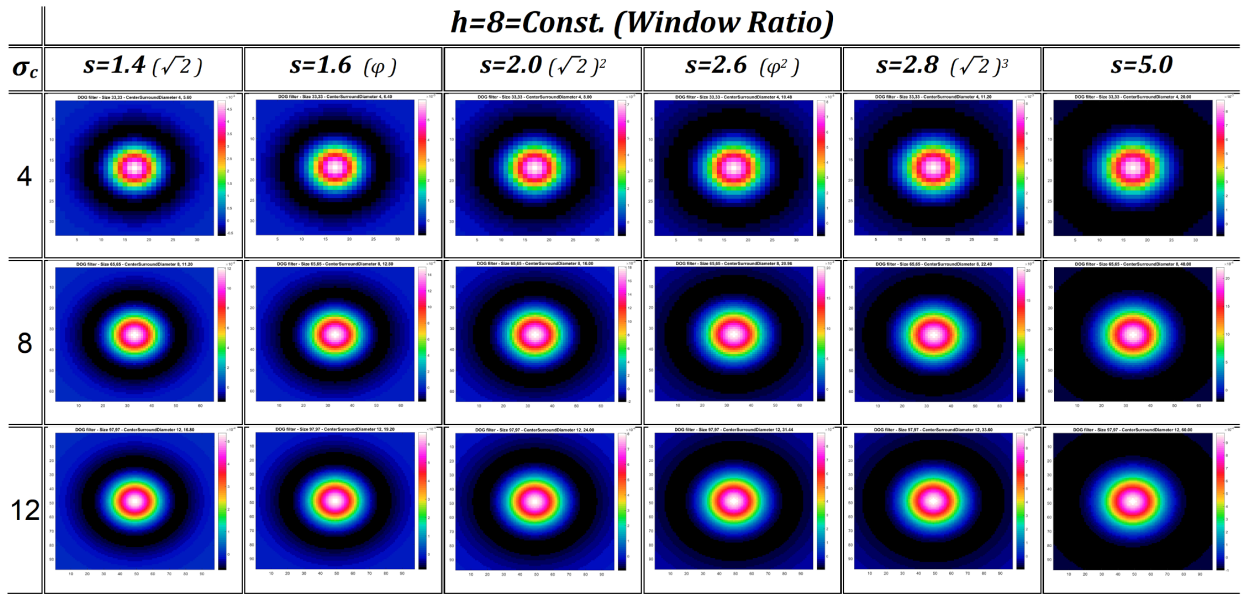

**Figure 2** DoG filters at three different scales ($\sigma_c$=4, 8, 12) of variations of *Surround ratio* ($s$=1.4, 1.6, 2.0, 2.6, 2.8, 5.0) with a constant *Window ratio* ($h$=8), presented in the jetwhite colormap. At each individual scale, by increasing the $s$ value (from *Left* to *Right*), we see a wider area of suppression effect caused by the surround Gaussian. As the scale increases, the overall filter size increases, resulting in a smoother DoG filter with higher computational cost. For the majority of experimental runs in our experiments, we have examined further $s$=1.6 and $s$=2.0. Modified from (Nematzadeh, 2018).

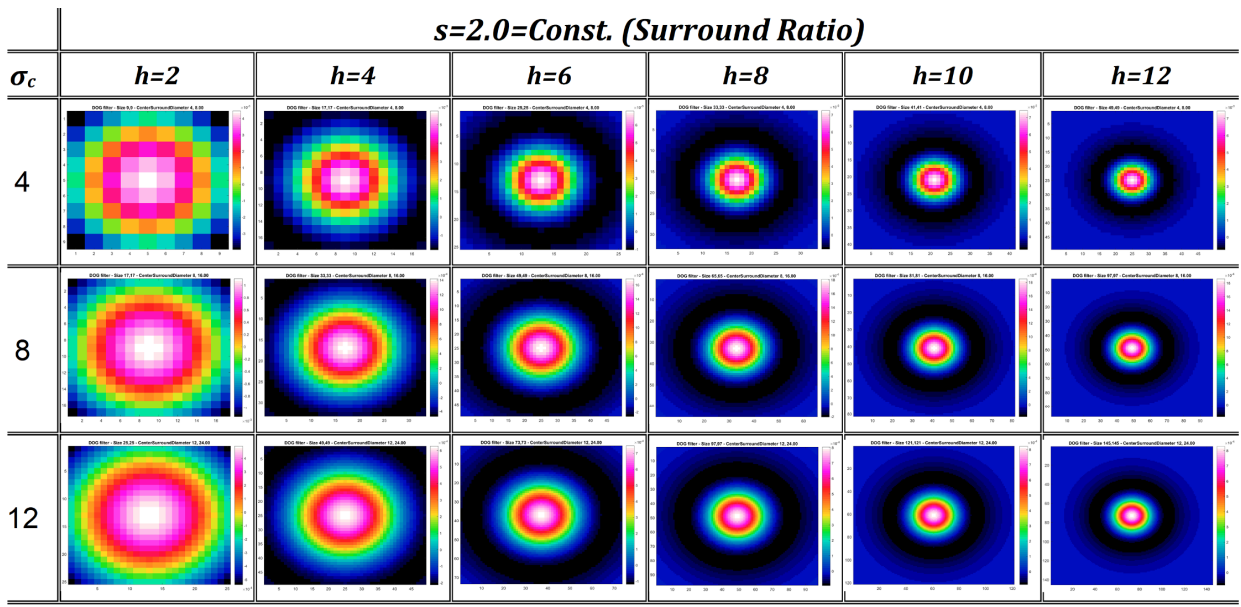

**Figure 3** DoG filters at three different scales ($\sigma_c$=4, 8, 12) of variations of *Window ratio* ($h$=2, 4, 6, 8, 10, 12) with a constant *Surround ratio* ($s$=2.0). Parameter $h$ determines how much of the center Gaussian and the surround Gaussian are included in the filter. Due to the constant value of *Surround ratio* ($s$=2), the only thing changing from *Top* to *Bottom* in each column is the window size, resulting in larger and smoother DoG filters. In each row, by increasing $h$ value, the window size also increases based on Eq. 3, and mainly what we see is more coverage of the center and surround Gaussians in the DoG filters by moving from *Left* to *Right*. *Modified from* (Nematzadeh, 2018).

It is noteworthy that despite the complexity and variety of retinal cell circuitry and coding, there are a few *constancy factors* common to them, valid even for amacrine and horizontal cells. The *constancy of integrated sensitivity* is an important factor noted in many of the abovementioned references such as (Enroth-Cugell and Shapley, 1973; Linsenmeier, Frishman, Jakiela et al., 1982; Croner and Kaplan, 1995). Croner and Kaplan (1995) found that "predicting changes in the response properties of ganglion cells across the retina on the basis of receptive field size may lead to erroneous conclusions" (p.23).

We investigate the effect of the *Surround ratio* ($s$) on the shape of the DoG filter in this section of the text. Variations of DoGs were tested with different values of $s$ ($1.4 \leq s \leq 5$) while keeping $\sigma_c$ and $h$ constant. $s$=1 lead to a Zero filter, since the center and surround Gaussians cancel each other. As indicated in Eq. (1) by increasing the value of $s$, we specify a wider area of suppression effect of the surround Gaussian, although its height declines due to normalization. We have examined $s$=1.4, 1.6 ($\approx \varphi$), 2.0, 2.6, 2.8, and 5.0 for the $s$ parameter while keeping other parameters of the model ($\sigma_c$ and $h$) constant. The DoG filters at three different scales ($\sigma_c$=4, 8, 12) and for $h$=8 have been presented in Figure 2 for these variations of $s$ parameter in our model.

As the DoG scale ($\sigma_c$) increases (from 4 to 12), both $\sigma_s$ and the window size increase (refer to Eq. (2) and Eq. (3) ). The title of each DoG filter in the figure indicates the window size as well as $\sigma_c$ and $\sigma_s$ of the filter. The results highlighted in Figure 2 indicates that the shape of the DoG filters is not very sensitive to the investigated modification range.

### 3.3. *Window ratio* ($h$) and DoG filters

We have investigated the effect of *Window ratio* ($h$) on the shape of the DoG filter in this section. As explained before, $h$ determines how much of the center Gaussian and the surround Gaussian are included in the filter. Variations of DoGs have been tested with different values of $h$ ($2 \leq h \leq 12$) while keeping $\sigma_c$ and $s$ constant. Two values for $s$ parameter have been considered ($s$=1.6 and $s$=2.0) to generate DoG edge maps (Figures 3 to 8) of sample patterns at multiple scales throughout this work.

The DoG filters with a constant *Surround ratio* of $s$=2.0 and at three different scales ($\sigma_c$=4, 8, 12) are presented for $h$=2, 4, 6, 8, 10, and 12 in Figure 3. As the figure shows, due to the constant value of *Surround ratio* ($s$=2), we see a change in the window size which increase from the left to the right column in the figure. The title of each DoG filter indicates the window size as well as $\sigma_c$ and $\sigma_s$ of the filter.

In particular, $h$=2 corresponds to the diameter of the center Gaussian, the part between the inflection points (68% center, 31% surround); $h$=4 corresponds to the diameter of outer Gaussian (95% center, 68% surround); and $h$=8 corresponds to the standard p<0.05 significance for the outer Gaussian (99.94% center, 95% surround). $h$=8 is used in experimental runs.

# 4. DoG representation of an image

## 4.1. DoG edge map at multiple scales

Applying a Gaussian filter to an image results in a blurred version of the image in our DoG model. Finding the Difference of two different scale Gaussian filters forms a corresponding scale of an edge map representation. A DoG representation of the image at multiple scales is referred to as an edge map at *multiple scales* or simply EMap-DoG for easy referral.

A sample DoG edge map of a Tile Illusion pattern is shown in Figure 4. A crop section (84×84px) of the Trampoline pattern (Kitaoka, 2000) was selected as an input image. The DoG scales are the standard deviation (sigma-$\sigma_c$) of the center Gaussian and in the figure the edge map is shown at five different scales: $\sigma_c$=0.5, 1.0, 1.5, 2.0 *and* 2.5. Other parameters of the model are the *Surround ratio* of 2 (*s*=2) and the *Window ratio* of 8 (*h*=8). The DoG filters in Figure 4 have been presented in the jetwhite colormap.

Scale-invariant processing, in general, is not sensitive to the exact parameter settings. The model's parameters ideally should be set in a way that at fine scales, they capture high-frequency texture details and at coarse scales, the kernel has appropriate size relative to the objects within the scene. We will explain how to define a proper set of values for parameters of Vis-CRF model ($\sigma_c$, *s, h, #of scales*) considering the pattern characteristics such as tiles and mortar sizes in the Café Wall stimulus in more detail in Section 5.

The multiple scale edge map representation of a cropped section of a Café Wall pattern with 200×200px tiles and 8px mortar are shown in Figures 5 and 6, as the output of the model, reveal the Twisted Cord elements along the mortar lines. This simple implementation of RGC responses based on DoGs, not only revealed the sharp edges when fine-scale filters are used but also by increasing the scale of the DoG ($\sigma_c$), other hidden information such as local texture information was revealed as well. Of those geometrical clues and cues of the pattern, our Vis-CRF model (Nematzadeh, Lewis and Powers, 2015; Nematzadeh and Powers, 2017a; Nematzadeh and Powers, 2016a; Nematzadeh, Powers and Lewis, 2016; Nematzadeh, Powers and Lewis, 2017) highlights the perception of divergence and convergence of mortar lines in the Café Wall illusion shown in Figures 5 and 6 and the appearance of tilt in Tile Illusions in general. We will show in the following how the model is capable of addressing the visual cues involved in the perception of illusory tilts such as the Café Wall illusion in Section 5, and some other Geometrical tilt illusions that will be explained in Section 6.

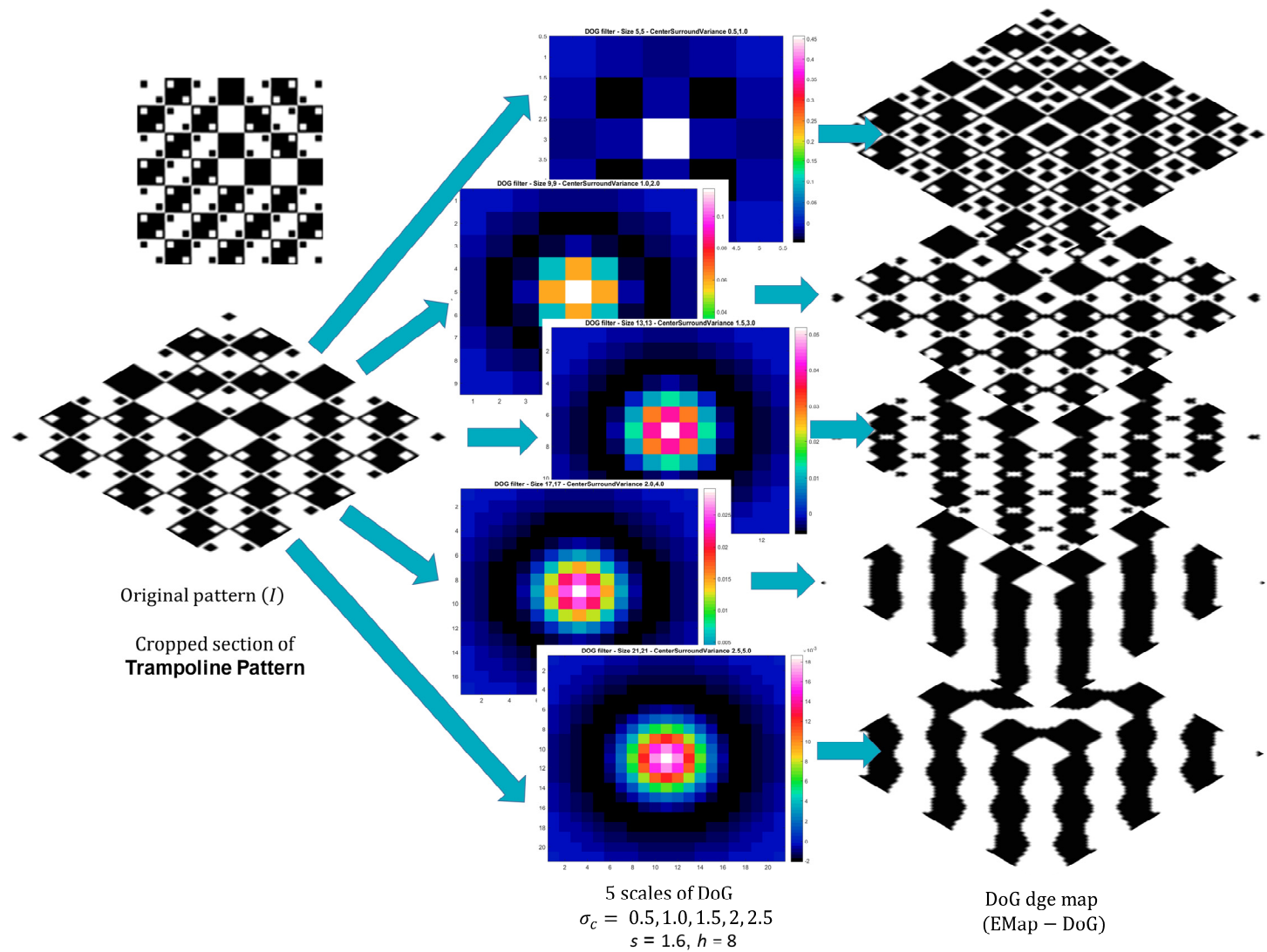

5 scales of DoG
$\sigma_c = 0.5, 1.0, 1.5, 2, 2.5$
$s = 1.6, h = 8$

DoG dge map
$(EMap - DoG)$

Original pattern ($I$)

Cropped section of
**Trampoline Pattern**

**Figure 4** A DoG edge map of a crop section of the 'Trampoline pattern' (Kitaoka, 2000 © *Used with permission from the designer*) as a sample of the tile illusion with the size of 84×84px. The scales of DoG filters are $\sigma_c$= 0.5, 1.0, 1.5, 2.0, 2.5. Other parameters of the model are $s$=2, and $h$=8 (*Surround* and *Window ratios* respectively). The edge map result on the left shows the extraction of different information from fine to coarse scales in the input pattern. We refer to this representation as a multiple scale DoG edge map in our model. 'Trampoline pattern' in high resolution can be found at: http://www.psy.ritsumei.ac.jp/~akitaoka/trampolineL.jpg (Nematzadeh, 2018).

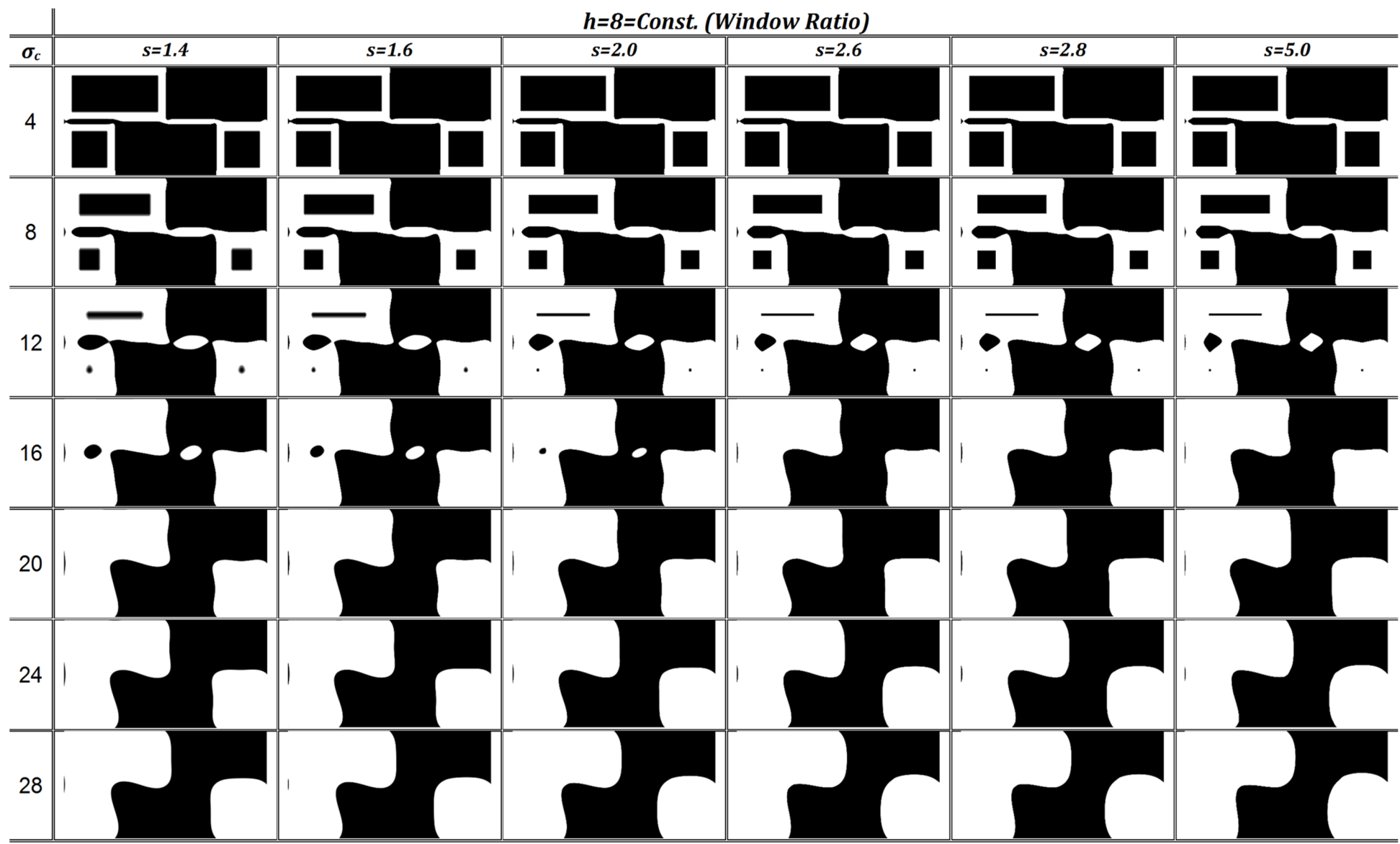

**Figure 5** DoG edge maps of a crop section of a Café Wall pattern with 200×200px tiles and 8px mortar of size $(T+M)×2T$ ($T$: tile size, $M$: mortar size), for variations of *Surround ratio* ($s$=1.4, 1.6, 2.0, 2.6, 2.8, 5.0) at a constant *Window ratio* ($h$=8), presented in Binary form at seven different scales ($\sigma_c$=4, 8, 12, 16, 20, 24, 28) (Nematzadeh, 2018).

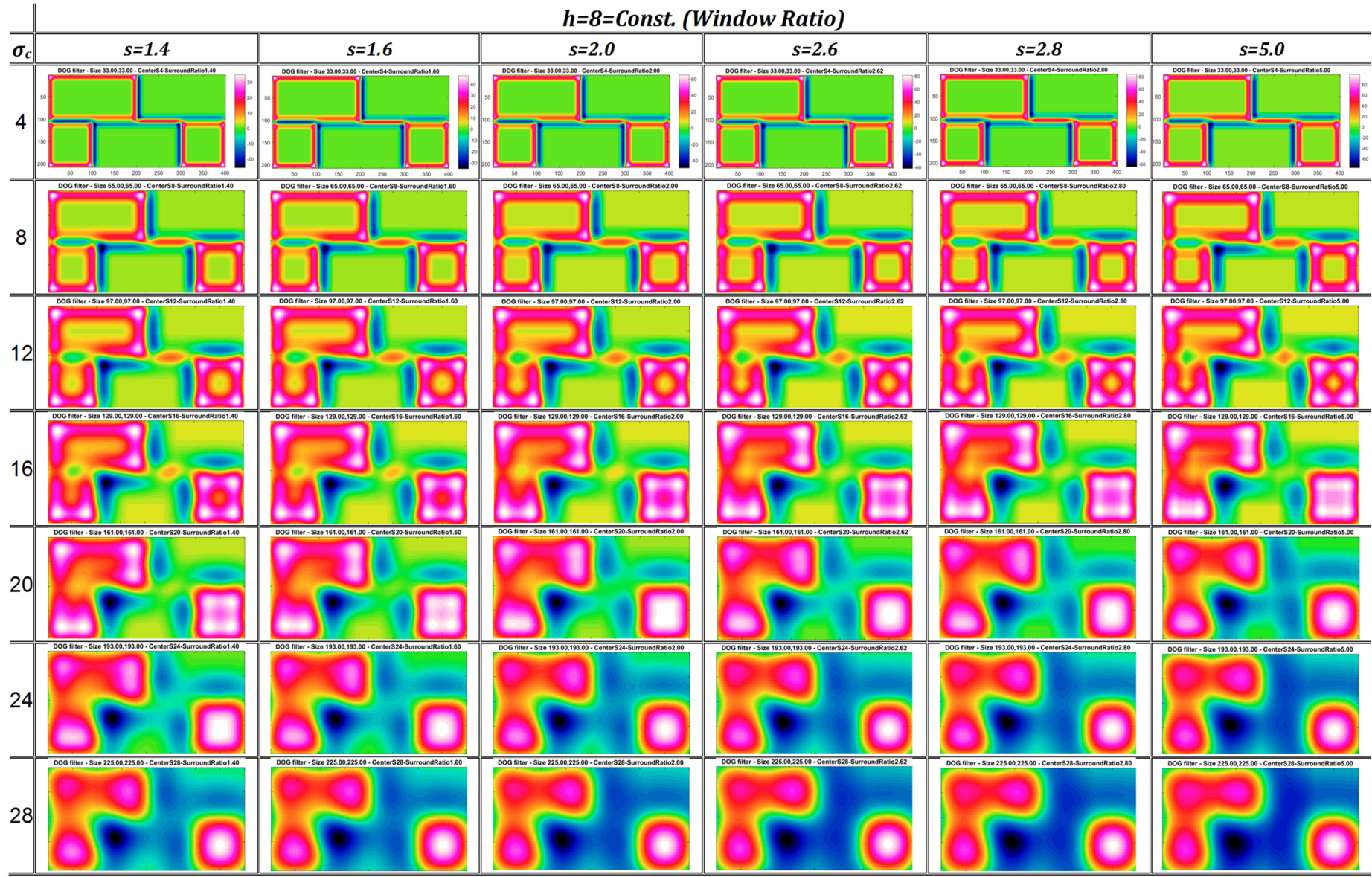

**Figure 6** DoG edge maps of a crop section of a Café Wall pattern with 200×200px tiles and 8px mortar of size ($T+M$)×2$T$ ($T$: tile size, $M$: mortar size), for variations of *Surround ratio* ($s$=1.4, 1.6, 2.0, 2.6, 2.8, 5.0) at a constant *Window ratio* ($h$=8), presented in the jetwhite colormap at seven different scales ($\sigma_c$=4, 8, 12, 16, 20, 24, 28) (Nematzadeh, 2018).

### 4.2. Surround ratio (s) and the DoG edge map

The DoG edge maps for the *Window ratio* of 8 (*h*=8) and variations of *s* parameter (*s*=1.4, 1.6, 2.0, 2.6, 2.8 and 5.0) have been given in Figures 5 and 6 for one cropped sample of a Café Wall pattern with 200×200px tiles and 8px mortar in Binary representation and the jetwhite colormap.

As the edge maps at multiple scales in Figures 5 and 6 show, there is not much difference at the finest scales ($\sigma_c$=4, 8) or very coarse scales ($\sigma_c$=24, 28) in this modification range of *Surround ratio* (*s*) presented for a constant *Window ratio* of *h*=8. The main difference in the edge maps is among the medium scales ($\sigma_c$=12, 16), which are the '*transient states*' from the grouping of tiles with mortar cues horizontally at fine scales, to separate the grouping of tiles in a zigzag vertical orientation at coarse scales, due to disconnection and disappearance of the mortar cues in the DoG edge maps at medium scales. The jetwhite false-colour representation (Powers, 2016) of the DoG edge maps presented in  (Figures 6 and 8) highlight the differences at medium scales more clearly.

It becomes clear from Figures 5 and 6 that the disconnection of mortar cues happens much quicker between scales for a higher range of *s* value. It is worth mentioning that despite the high variations of the surround Gaussians, there is not a drastic difference between DoG filters and the output edge maps within these ranges of variations. We have chosen either *s*=2 or *s*=1.6 (Earle and Maskell, 1993; Marr and Hildreth, 1980) in the majority of our experiments in this work and in our previous work for general investigation on Geometrical illusions (Nematzadeh, Powers and Lewis, 2019; Nematzadeh, Powers and Lewis, 2017).

### 4.3. Window ratio (h) and the DoG edge map

As shown in Figure 1 (Bottom-right) for a constant *Surround ratio* (*s*=2), *h*=2 corresponds to the diameter of center Gaussian, the parts between the inflection points, which is 68% center and 31% surround; *h*=4 corresponds to the diameter of surround Gaussian between inflection points, 95% center and 68% surround; *h*=8 corresponds to DoG filter with 95% surround; and *h*=12 corresponds to more than 99.6% of the significance to be included in the surround Gaussian.

The DoG edge maps for variations of *h* parameter (*h*=2, 4, 6, 8, 10, 12) have been provided in Figures 7 and 8 for *s*=1.6, for a crop sample of a Café Wall pattern with 200×200px tiles and 8px mortar in Binary representation and in the jetwhite colormap. The size of the cropped section is (*T+M*)×*2T* (*T*: tile size, *M*: mortar size). There is not a significant difference between the DoG edge maps at *s*=1.6 and *s*=2.0 (Nematzadeh, 2018).

The precision of DoG outputs are not quite satisfactory at *h*=2 and *h*=4 because these filters are not smooth, but the convolved DoG result is quite smooth with a high computational cost when *h*=12. A feasible and reliable range for *Window ratio* is between *h*=6 to *h*=10. In this range, *h*=8 with less than 5% of the insignificance to lie outside the surround Gaussian is a good empirical value for this parameter in our model. Technically, if we see a significant change as we increase $h$ by 2, then $h$ is too small. Also, the window size does not affect the output once the surround approaches zero.

Comparing the DoG edge maps at variations of $h$ parameter (Figures 7 and 8), we see that for $h=2$, the cues of the mortar lines still exist even at the coarsest scale ($\sigma_c=28$). At fine to medium scales when $h$ increases, the persistence of mortar cues becomes weaker. However, much of the information at different scales of the edge maps for $h=2$ and $h=4$ are redundant. If we want to capture all the pattern features (mortar lines and tiles in this stimulus) in one edge map representation, the number of scales should go beyond the seven scales defined in our investigations at the used range of $h$ value.

Marr and Hildreth in their theory of edge detection noted that: "Another, more practical, reason why 'edge detecting' operators have previously less than optimally successful in computer vision is that most current operators examine only a very small part of the image, their 'receptive fields' are of order of 10 to 20 image points at most. This contrasts sharply with the smallest of Wilson's four psychophysical channels, the receptive field of which must cover over 500 foveal cones." (Marr and Hildreth, 1980; p.200).

Also there are biological correlates in our vision for the existence of a big surround (inhibition) when using the DoG modelling for the lateral inhibition of simple cells. Bekesy in his book of 'sensory inhibition' noted that "in vision, the neural unit has a wide inhibitory area, so that the lateral inhibition effects probably extend to larger regions. It is possible however that the lateral speed of the neural unit is different at different neural levels" (Bekesy Von, 1967; p. 228) (check Fig. 68; p. 79 for more detail where $S$ in the figure is showing sensory area, and $R$ is the refractory area/inhibition; $S/R=1.6$ in eye and 1.2 in skin). About the relative size of the surround to the center, Linsenmeier et al. (1982), noted in his review of other researcher that "Cleland et al. (1973) found that the surround diameter was, on the average 5.13 and 4.25 times larger than the center diameter in sustained and transient cells respectively; Enroth-Cugell and Robson (1966) found a value of 6.74 for their X cells; and in the LGN So and Shapley (1981) have reported values of 4.58 for X and 7.99 for a small sample of Y cells" (p. 1182). He then continues by stating that: "Our value for X cells is 3.88, which is not too different, but for Y cells we found an average of only 1.49. This is probably due in part to the tendency to underestimate surround size (p, 1177) and may also result from a rather small number of points at very low spatial frequencies in Y cells. There is, however, some independent evidence that $r_s/r_c$ is smaller in Y than in X cells" (p. 1182).

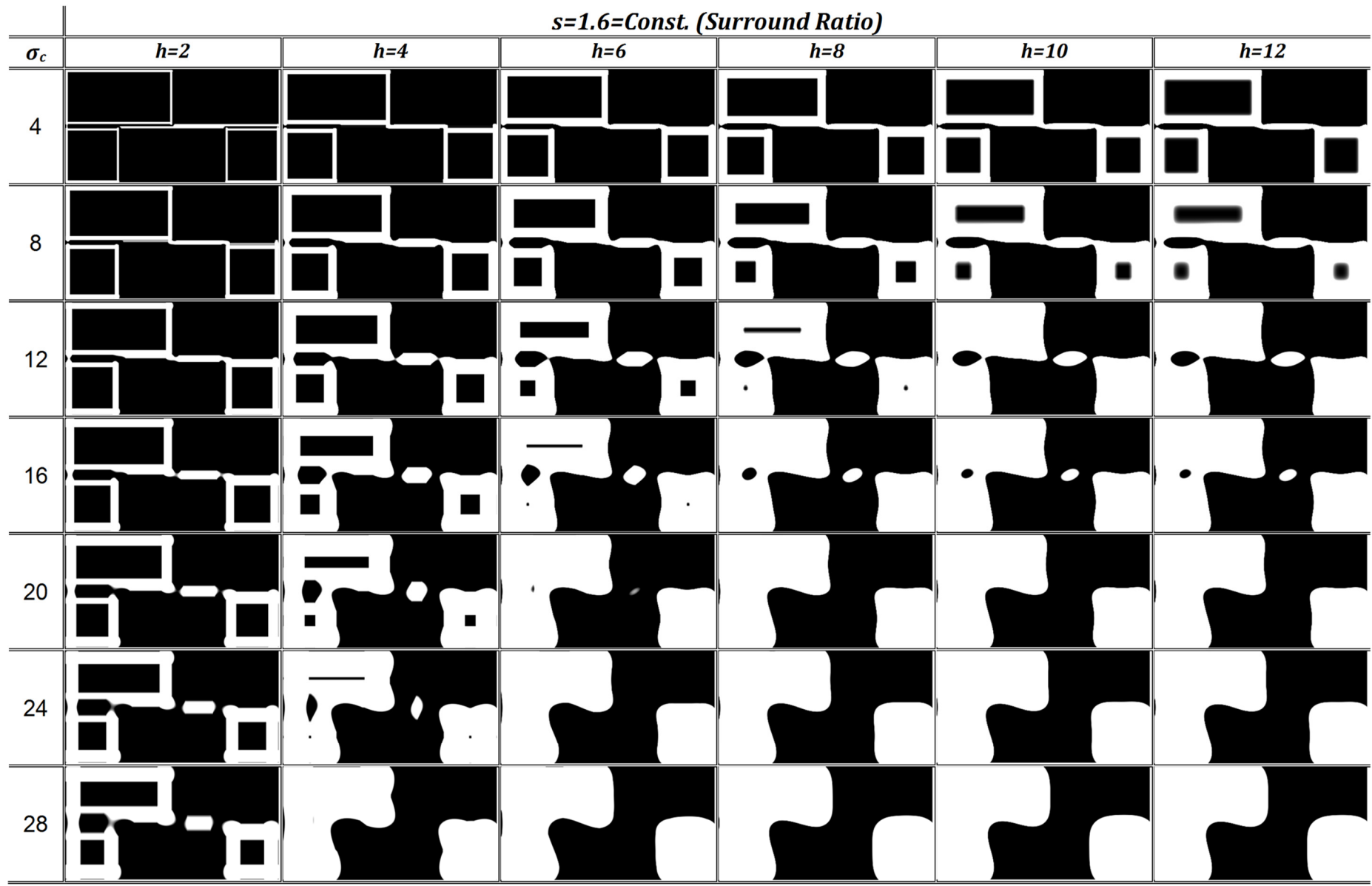

**Figure 7** DoG edge maps of a crop section of a Café Wall pattern with 200×200px tiles and 8px mortar of size $(T+M)×2T$ ($T$: tile size, $M$: mortar size), for variations of *Window ratio* ($h$=2, 4, 6, 8, 10, 12) at a constant *Surround ratio* ($s$=1.6), presented in Binary form at seven different scales ($\sigma_c$=4, 8, 12, 16, 20, 24, 28) (Nematzadeh, 2018).

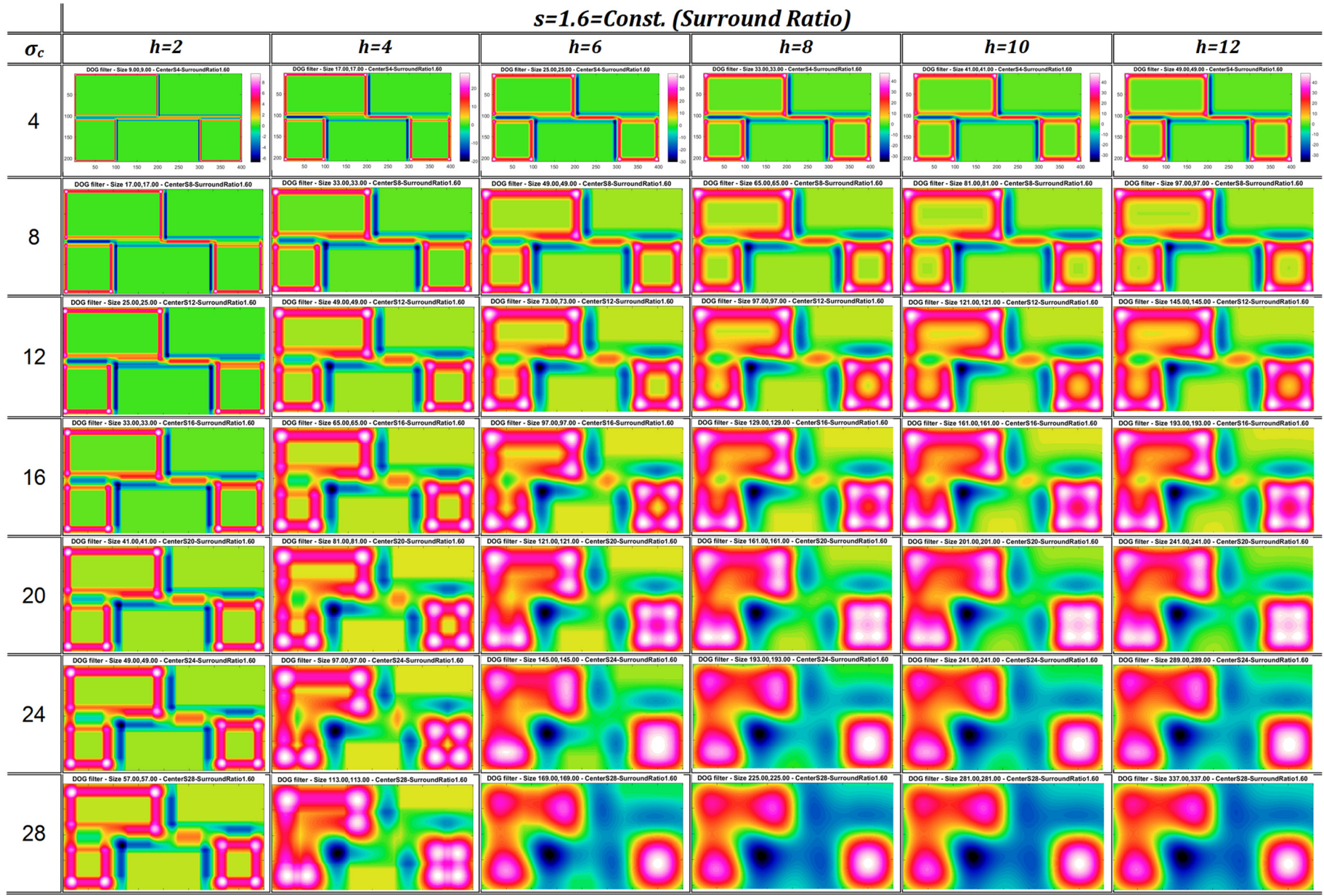

**Figure 8** DoG edge maps of a crop section of a Café Wall pattern with 200×200px tiles and 8px mortar of size ($T$+$M$)×2$T$ ($T$: tile size, $M$: mortar size), for variations of *Window ratio* ($h$=2, 4, 6, 8, 10, 12) at a constant *Surround ratio* ($s$=1.6), presented in the jetwhite colormap at seven different scales ($\sigma_c$=4, 8, 12, 16, 20, 24, 28) (Nematzadeh, 2018).

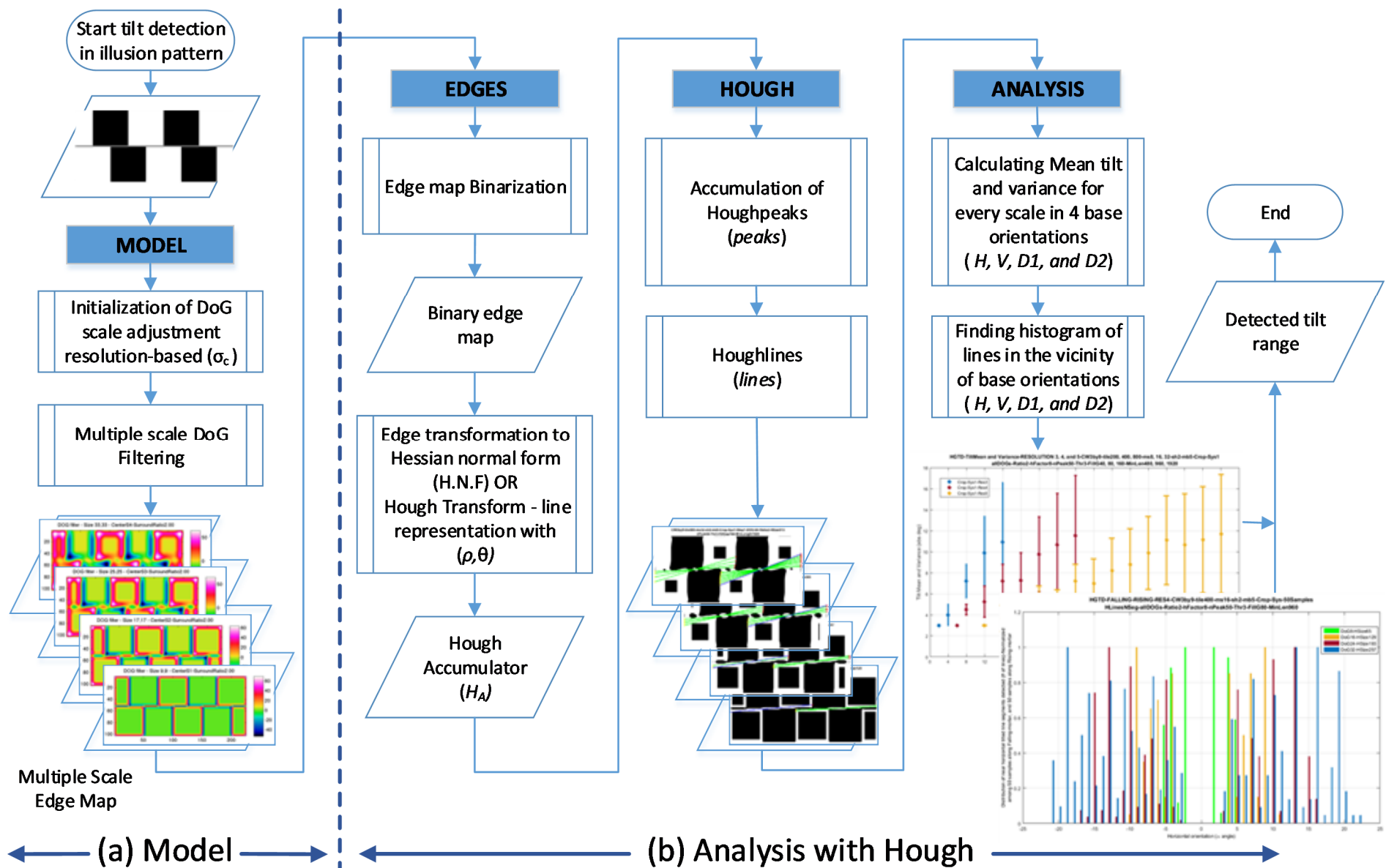

**Figure 9 Flowchart of experimental process** (Nematzadeh, 2018)

Comparing the edge maps (in Figures 7 and 8) for $h$=2 and other higher values of $h$ parameter, we see that $h$=8, 10, and 12 can detect both mortar lines and tiles information in a parsimonious manner at seven different scales ($\sigma_c$=4, 8, 12, 16, 20, 24, 28). We also need a gradual change of scales especially at fine to medium scales for detecting illusory cues in the Tilt Illusion patterns. Also needed is a set of DoG filters where their scales are adjusted based on the pattern characteristics (the mortar and tile sizes in the Café Wall). We demonstrated in details in (Nematzadeh, 2018; Nematzadeh and Powers, 2017c; Nematzadeh, Powers and Lewis, 2017; Nematzadeh and Powers, 2016b), how this gradual change of scales is necessary in our DoG model (Vis-CRF) for a reliable edge representation of the pattern, and how it contributes to the "persistence of mortar cues" in the edge map. We have shown how the persistency of mortar cues in the edge map (EMap-DoG) indicates the strength of tilt effect in variations of the Café Wall pattern in (Nematzadeh and Powers, 2017c). In the majority of the experimental runs, $h$=8 has been chosen for the *Window ratio*, which corresponds to the traditional loss of 5% of the surround, keeping 95% of the outer Gaussian.

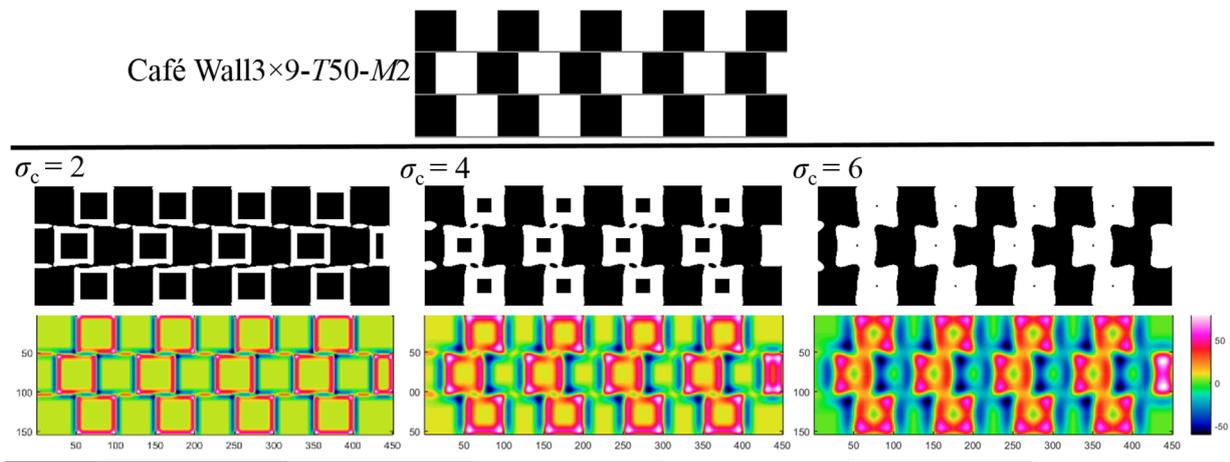

**Figure 10** *Top*: A Café Wall pattern of 3×9 tiles with 50×50px tiles and 2px mortar. *Middle*: Binary representation of the multiple scale DoG edge map at three different scales ($\sigma_c$) ranges from 2 to 6. The other two parameters of the DoG model are constant: *s = 2, h = 8* (*Surround* and *Window ratios* respectively). *Bottom*: The jetwhite colormap representation of the multiple scale edge map (Nematzadeh, 2018).

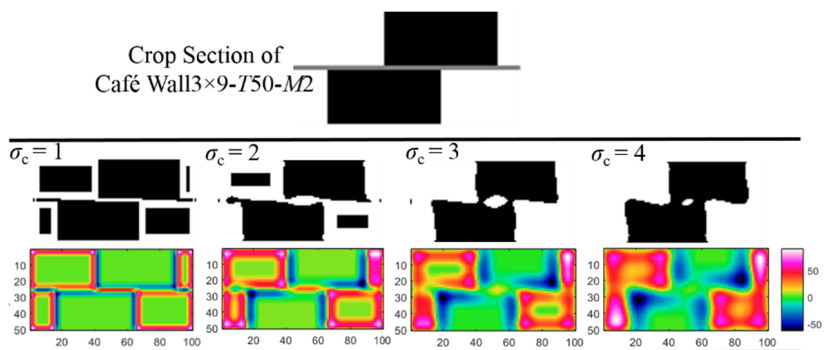

**Figure 11** *Top*: A crop section with a single slanted line segment from a Café Wall of 3×9 tiles with 50×50px tiles and 2px mortar (Café Wall3×9-T50-M2). *Middle*: Binary representation of the multiple scale DoG edge map at four different scales ($\sigma_c$) ranges from 1 to 4 (*Left* to *Right*), *s = 2, h = 8* are constant in the DoG model here. *Bottom*: The jetwhite colormap representation of the DoG edge map (The range of scales is chosen in such a way as to capture the near horizontal tilt cues in the Café Wall pattern: $\sigma_c$=0.5M to 2M here) (Nematzadeh, 2018).

## 5. Model and Image processing pipeline

We now discuss the details of an image processing pipeline used to extract edges and how to analyse their angles of tilt, as shown in Figure 9 for a cropped section of a Café Wall pattern of the size *2×4.5* tiles (the precise height of cropped section is *2Tiles+Mortar=2T+M*). The processing explanations concentrated on the analysis of the tilt effect in the Café Wall illusion, to include every detail used, to simulate and quantify the tilt angle in this stimulus can be found in (Nematzadeh, 2018-Chapter 4).

### 5.1. MODEL

The DoG representation at multiple scales as the output of modelling ON-center OFF-surround activation of retinal ganglion cells (RGCs) at different scales is presented in Figure 10 for the Café Wall of $3×9$ tiles with $50×50$px tiles ($T$) and 2px mortar ($M$) (Café Wall 3×9-T50-M2). For convenience, we focus on small cropped sections of the original image and this is illustrated in Figure 11, with its DoG representation for a different range of $\sigma_c$, varied from 1 to 4. The crop is selected from the Café Wall of 3×9 tiles described above. The most fundamental parameter in the model is the diameter of the center Gaussian ($\sigma_c$), and the point of maximum illusory effect occurs when $\sigma_c$ matches the pattern features and their characteristics, for the Café Wall, specifically the mortar size ($M$). To extract the tilted line segments along the mortar lines, we thus set $\sigma_c$ to be of the same order as the mortar size. We undertook empirical test simulations to find an appropriate range for $\sigma_c$, testing range of *1/2M* to *2M* with incremental steps of *1/2M*. Others use similar DoG representations in the deepest layer of a DNN (Lv, Jiang, Yu et al., 2015).

The DoG outputs in Figures 10 and 11 show that by increasing the scale, the corner effect is highlighted. This effect with the appearance of tilted line segments in the edge map has an effect on the appearance of square tiles which look similar to trapezoids, inducing convergent and divergent mortar lines. The trapezoid shape has been referred to as wedges in some previous studies on the Café Wall pattern and declared to be the explanation for the tilt effect in the Café Wall illusion (Gregory and Heard, 1979). On the other hand, the output of lower scale DoG filters preserves the outline and shape of the tiles and the connectivity of them by the mortar line. This local connectivity of tiles in adjacent rows by the mortar line tends to become disconnected when the scale of the filter is increased. So as the DoG response indicates, by a moderate change of scale, we can see two '*incompatible groupings*' of pattern elements. The grouping of tiles in two consecutive rows by the mortar lines at fine scales with nearly horizontal orientation, and then by increasing the scale, we see a grouping of tiles in a zigzag vertical direction. We believe that these two groupings of pattern elements that occur simultaneously at multiple scales, which are not consistent in their orientations, cause the illusory tilt. These two incompatible groupings, along with systematic differences relating to the relative size of Gaussian and pattern scales, result in illusory tilt effects that reflect changes in size and density with eccentricity, predicting the change in illusion effects according to distances from the focal point in the pattern versus distance in the retinal image from the fovea to the periphery.

## 5.2. Introduction to the Tilt analysis pipeline

The DoG transformation reveals the tilt cues in Tilt/Tile Illusions, and the quantitative measurement of tilt angles lets us compare them with the tilt perceived by a human observer. For this, we embed the DoG model in a processing pipeline involving multiple standard image processing transformations. As the focus in this work, we explained the details of DoG processing and the edge map representations of the patterns in our model. Let us briefly explain the tilt analysis designed in our investigations in this Section (The details of tilt analysis is provided in (Nematzadeh, 2018-Chapter 4)).

For the analytical approach presented in the right half of the flowchart-Figure 9, we have used Hough space to extract tilt angles of the detected line segments in the edge maps. This analytical approach can be replaced by **any other application-based method** to extract the *relevant features* necessary from the edge map for further investigations (the edge map is a DoG filtered response at multiple scales and is actually the encoded information of the contrast inside an image). Figure 9 shows the flowchart of our model and the analytical tilt processing in the right half that contains three stages of EDGES, HOUGH and ANALYSIS. The output of the MODEL is an edge map at multiple scales that is sent to these three Hough stages for further *analysis of tilt*. After applying the Hough Transform, the edge information is mapped to a Hough space. We then are able to extract the angle of each line segment that is detected from the edge map. By applying some constraints to the Hough algorithm and elimination of the line segments, we can capture the information of the proper set of line segments for further analysis across the multiple scales. In the final stage, we find the contributions of line segments detected around the predefined reference orientations (along horizontals, verticals and diagonals in our investigations). Further details of analytical tilt extraction and quantitative measurement of tilt angles are provided in Chapter 4 of (Nematzadeh, 2018).

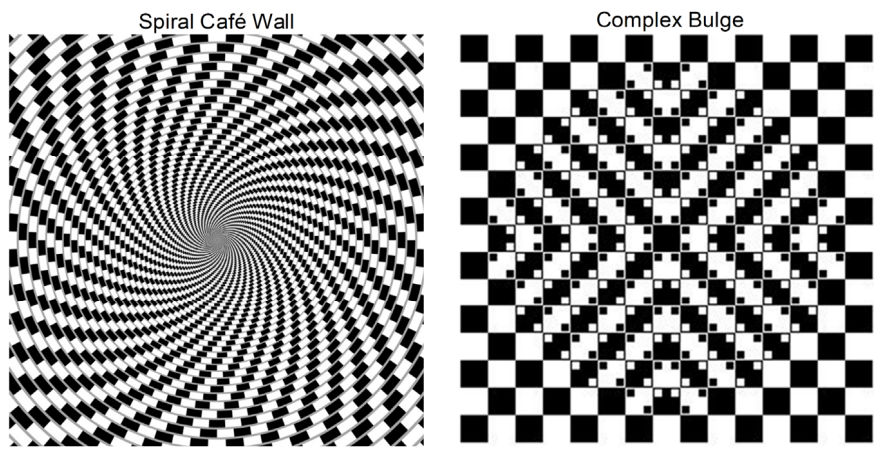

**Figure 12.** Two complex Tile illusions: (Left) Spiral Café Wall (Kitaoka, 2007), (Right) Complex Bulge pattern (Kitaoka, 1998).

# 6. Other Tilt/Tile illusions

In this section, we are going to show the generalization of Vis-CRF model by addressing a broader range of Tile/Tilt illusions, and by investigating two complex illusions shown in Figure 12 (Spiral Café Wall and Complex Bulge patterns). We briefly illustrate the results of Hough analysis stages for these two illusions as well. It is noteworthy to say that for detection of tilt angles in these types of illusions, we need to investigate local tilts at different parts of the pattern (around the focal point – Local investigation of tilt). Vis-CRF model can reliably predict these local tilt cues as the underlying mechanism in the illusory tilts we perceive but we need to validate the predictions of our model by performing psychophysical experiments which is a priority in our future work.

## 6.1. EMap-DoG

Figure 13 shows the edge map representations for the Spiral Café Wall illusion with the size of 875×875px. The edge maps are presented at twelve different scales ($\sigma_c$ = 1 to 12) with incremental steps of 1 in the jetwhite colormap on top and in binary form at the bottom half of the figure. The pattern is among the complex Tile Illusion patterns compared to the Café Wall illusion. Although in this pattern the rows of Café Wall tiles are arranged in a circular way with mortar lines in between (like rings of tiles), we get an impression of their arrangement as being in a spiral organization that is an illusory percept of the pattern. Due to the circular design of the pattern, the size of tiles and mortar lines has been increased from the center to the surrounding region of the pattern and encoding of tiles as well as mortar lines are different from the original Café Wall pattern which consists of constant tile size and mortar size. The appearance of a complete tile in the edge map is dependent on the scale of the edge map relative to each individual tile size in a ring of tiles, this also holds true for the appearance of mortar lines.

What we see at fine scales of the edge map is the extraction of fine details from mortar lines to tiles edges in the pattern. The tiles are connected through mortar lines at very fine scales ($\sigma_c$ = 1, 2). As the scale increases, we see some blending of colour in the DoG output and that the mortar cues start to disconnect. As a result, the grouping of tiles by the mortar lines disappears and a different grouping of tiles starts to be revealed at scale 4 ($\sigma_c$ = 4) for the central tiles, and near scale 7 ($\sigma_c$ = 7) for the peripheral ones close to the outer border of the pattern. The new grouping of tiles generates curved lines, from the center of the pattern to the periphery, which is thin at the center and thick at the other end. Similar groups appear around the whole pattern, moving from the center to the outer region with a slight circular rotation of a similar curved line at mid to coarse scales in the edge map. Another thing worth mentioning is the size extension of the central hole in the DoG edge map as the scale increases. All of these cues can contribute to the perception of the Spiral Café Wall rather than the circular arrangement of the Café Wall tiles in the pattern.

The next Geometrical Illusion investigated here is the Complex Bulge pattern (Kitaoka, 1998). This pattern consists of a simple checkerboard background and some superimposed White/Black dots on Black/White tiles, arranged in the center of the checkerboard, giving the impression of a central bulge in the pattern. We refer to these tilt patterns as second-order tilt effect illusions, and the superimposed dots on their backgrounds give some impression of foreground-background percept. Changing the positions of dots on the textured (Tiled) background results in some tilt, wave or bow effects along the edges as well as expansion and/or contractions on checkers corners (Nematzadeh, Lewis and Powers, 2015). More examples of these type of illusions are provided in section 6.3, for the patterns generated by the lead author of this work. The DoG edge maps at multiple scales for the Complex Bulge pattern are presented in Figure 14 in the jetwhite colormap (top) and in binary form (bottom) at twelve different scales (from $\sigma_c = 1$ to $12$ with the incremental steps of $1$). The pattern investigated has a size of $574 \times 572$px. At fine scales ($\sigma_c = 1$, $2$), the DoG output reveals the fine details of the pattern including the edges of tiles and the superimposed dots. What we see at fine scales is a grouping of tiles with superimposed dots in a circular arrangement around the center.

The impression of central bulge in the edge map lasts till nearly scale 5 ($\sigma_c = 5$), and at this scale, we see a transient state from the central bulge effect as the result of the grouping of tiles with superimposed dots to a different grouping of tiles in an X shape organization of identically coloured tiles in the pattern. This grouping of tile elements persists from mid to coarse scales when the cues of fine-scale superimposed dots have been disappeared in the edge map.

Again, what we believe as the explanation of the illusory percept of the central bulge in the pattern is a simultaneous sampling of the pattern elements at multiple scales in the retinal encoding of the pattern, which results in two incompatible groupings of pattern elements which contribute to the illusory perception of the central bulge in the pattern. Based on the relative size of superimposed dots and the tiles of the checkerboard, the impression of central bulge is very clear for the given pattern. It is obvious that by decreasing the size of superimposed dots the persistency of dot cues at fine scales of the DoG edge map would be reduced, and we encounter with a weaker illusory bulge effect at the center.

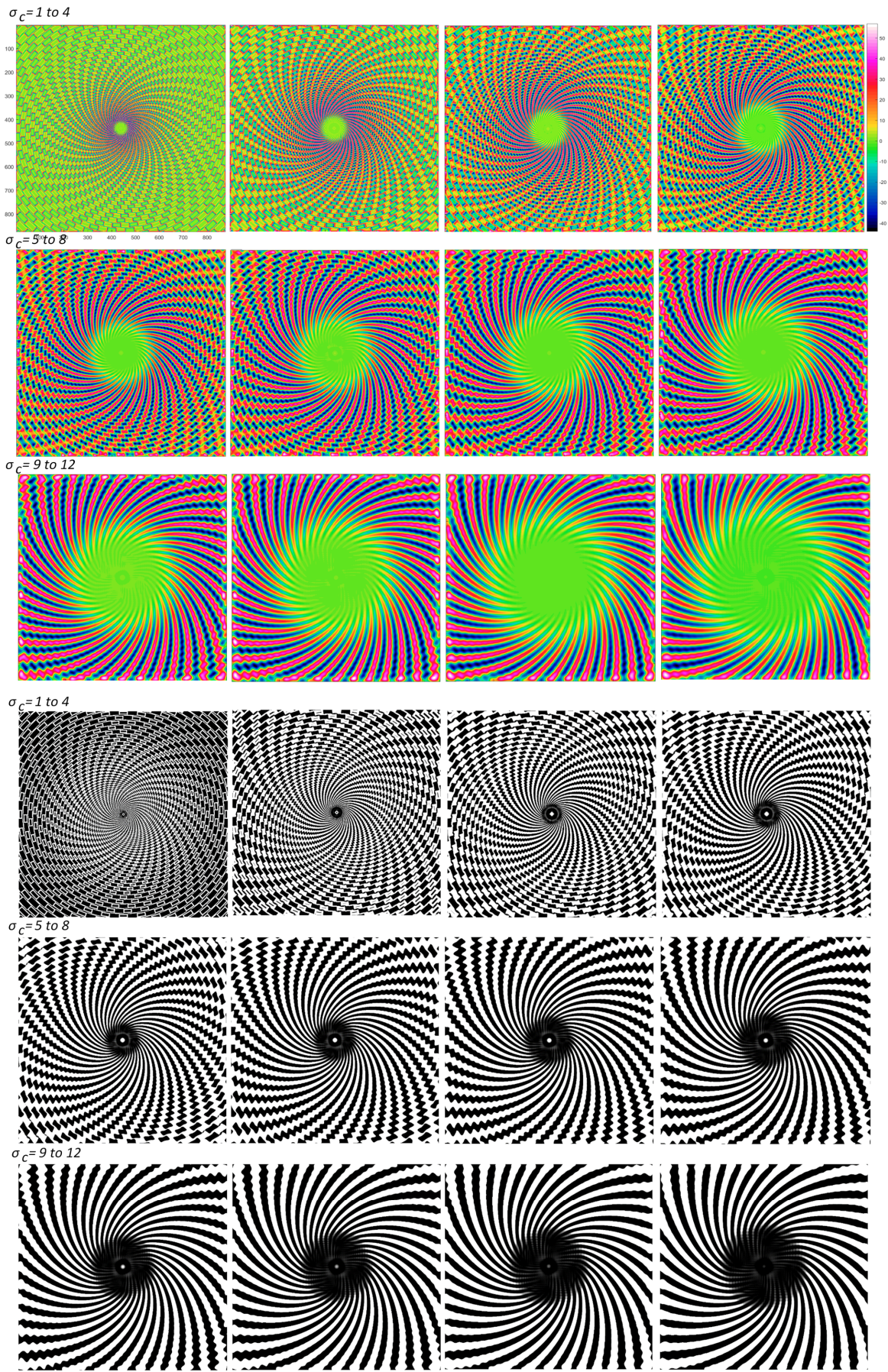

**Figure 13** DoG edge maps at twelve different scales ($\sigma_c$= 1 to 12) from *Left* to *Right* and *Top* to *Bottom*, with incremental steps of 1, presented in the jetwhite colormap (on *Top*), and in binary form (at the *Bottom*) for the Spiral Café Wall illusion. The investigated pattern has a size of 875×875px. The fundamental parameters of the model are *s*=1.6, *h*=8 (*Surround* and *Window ratios respectively*). Modified from (Nematzadeh, 2018).

$\sigma_c$ = 1 to 4

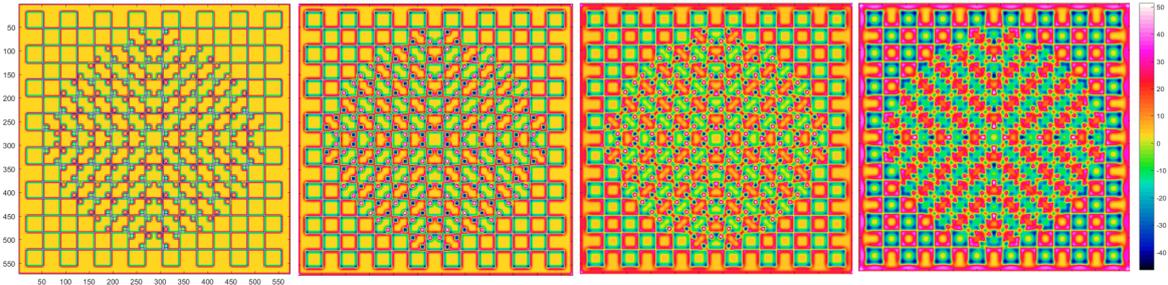

$\sigma_c$ = 5 to 8

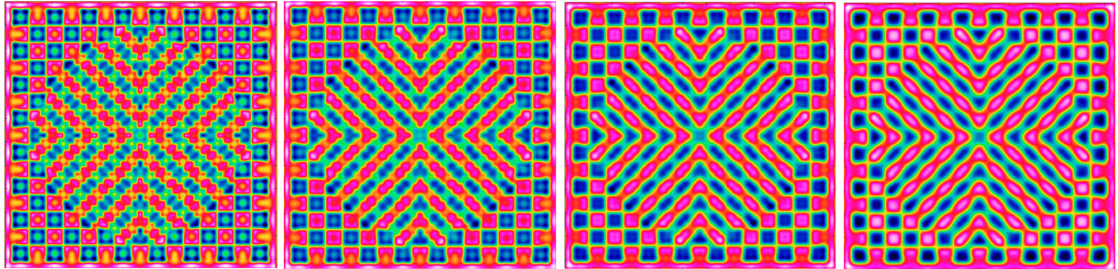

$\sigma_c$ = 9 to 12

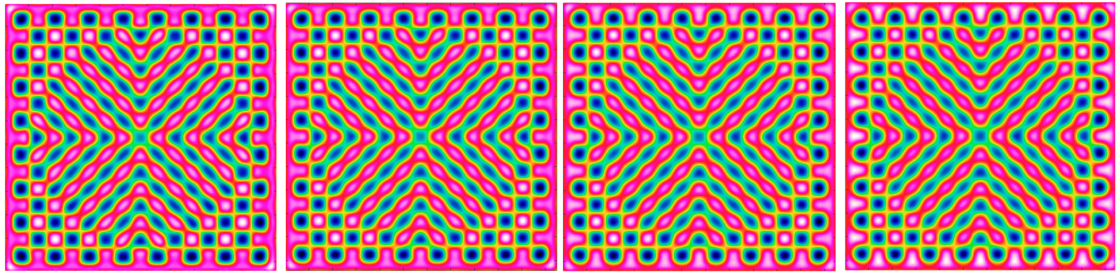

$\sigma_c$ = 1 to 4

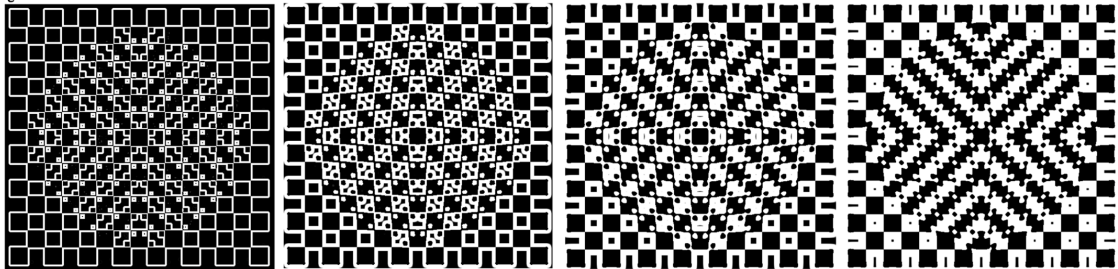

$\sigma_c$ = 5 to 8

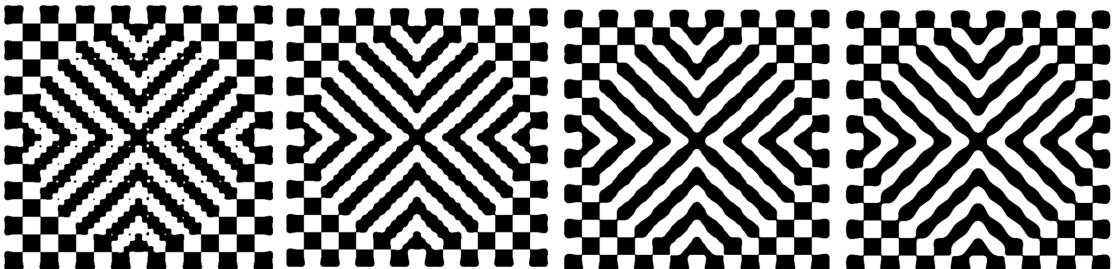

$\sigma_c$ = 9 to 12

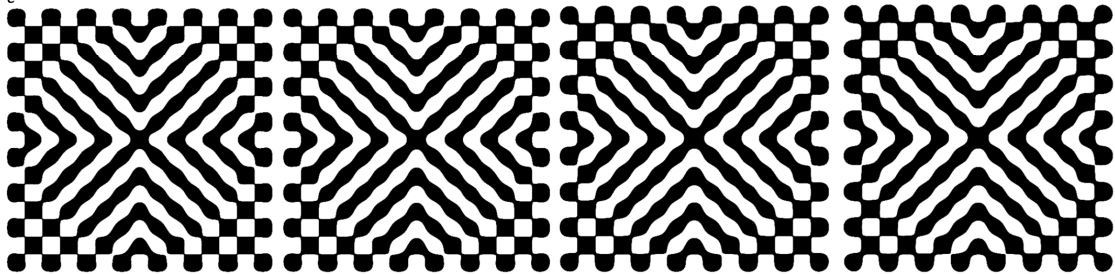

**Figure 14** DoG edge maps at twelve different scales ($\sigma_c$= 1 to 12) from *Left* to *Right* and *Top* to *Bottom*, with incremental steps of 1, presented in the jetwhite colormap (on *Top*), and in binary form (at the *Bottom*). The investigated pattern has a size of 574×572px. The fundamental parameters of the model are $s$=1.6, $h$=8 (*Surround* and *Window ratios respectively*). Modified from (Nematzadeh, 2018).

## 6.2.  Detecting tilts in the edge maps

Our Vis-CRF model provides new insights into physiological models (Carandini, 2004b; Field and Chichilnisky, 2007; Gollisch and Meister, 2010) as well as supporting Marr's theory of low-level vision (Marr and Hildreth, 1980; Marr and Ullman, 1981). In this section we briefly illustrate how we further analyse the detected tilted lines in the DoG edge maps of the investigated patterns at multiple scales (presented in Figures 13 and 14) using the Hough analysis pipeline, briefly explained in Section 5.2 with more details in (Nematzadeh, 2018-Chapter 4; Section 4.4).

The model's early stage output is investigated to quantify the degree of tilt using the Hough Transform in place of the later higher-order cortical processing. Mean tilt and standard deviation of the detected tilt in line segments are calculated for every scale of the edge map, providing quantified predictions for these experiments with human subjects.

Two output results of the HOUGH stage are presented in Figure 15, as detected houghlines shown in Green, displayed on the binary edge maps at multiple scales of the DoGs. The parameters of the model to generate the edge maps are $s$=1.6 and $h$=8 for the *Surround* and *Window ratios* respectively here. The DoG scales ($\sigma_c$) are presented on the Left corner of each scale of the edge maps in the figure.

On the top of Figure 15 , the detected houghlines are shown in Green, displayed on the DoG edge map of the Spiral Café Wall illusion at twelve different scales (from $\sigma_c$= 0.5 to 5.5 with incremental steps of 0.5). The hough parameters for detecting tilt angles are provided in the figure caption. At fine scales ($\sigma_c$= 0.5 to 1.5), we see small detected line segments connecting the tiles' edges with the mortar lines in a circular manner around the center and since the edges are dense at the central part of the pattern, the hough algorithm detects other lines with another direction from the center to the outer region, concentrated in the central part of the pattern at the finest scale ($\sigma_c$= 0.5) which extends to the surround region as the scale increases. At scales 1.5 and 2.0 ($\sigma_c$= 1.5, 2.0), we see similar distributions of these two different lines with a circular organization around the center (connecting tiles with mortar lines) as well as small line segments whose integrations result in some kind of curved lines from the center to the periphery of the pattern, connecting tiles together in the regions where the mortar cues are disappeared in the edge map.

At medium scales ($\sigma_c$= 2.0 to 3.5)-second row of the edge map (with modified hough parameters), we can see these lines more clearly, showing how the majority of houghlines are detected along the center to the periphery with their integration resulting in construction of curved lines from the center to the surround region of the pattern (concentrated on the central regions and up to the middle of the pattern) where the mortar cues start to fade. In the outer region of the pattern, when the mortar cues still exist in the edge map, the houghlines are detected connecting tiles with the mortar lines in nearly circular arrangements. Also we can see the increasing size of the central hole in the edge map of the pattern by increasing the DoG scale. At coarse scales ($\sigma_c$= 5, 5.5)-third row of the edge map, the integrated curved lines are extended to the border of the pattern.

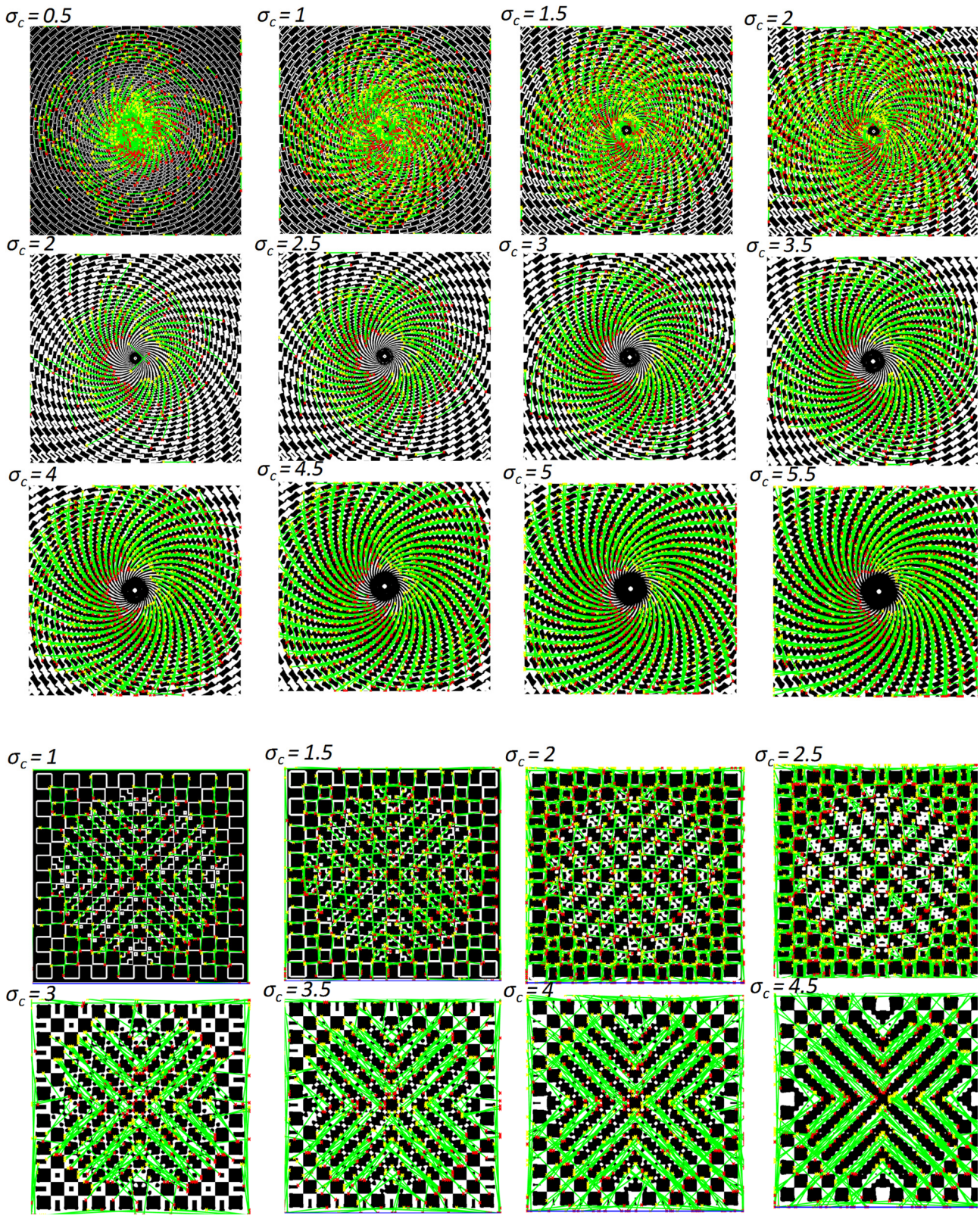

**Figure 15** *Top*: Edge/tilt detection of the edge map at multiple scales for the Spiral Café Wall illusion ($\sigma_c$= 0.5 to 5.5 with incremental steps of 0.5) with hough parameters of *FillGap*=5, *MinLenght*=50 (1st row) and *FillGap*=5, *MinLenght*=100 (2nd and 3rd Row). *Middle*: Edge/tilt detection of multiple scale edge map of the Complex Bulge pattern ($\sigma_c$= 1 to 4.5 with incremental steps of 0.5) with hough parameters of *FillGap*=5, *MinLenght*=50 (1st row) and *FillGap*=10, *MinLenght*=100 (2nd row; the explanation for Hough parameters can be found in (Nematzadeh, 2018)). Crosses mark start (yellow) and end (red). Other fundamental parameters of the model are $s$=1.6, and $h$=8 (*Surround* and *Window ratios* respectively) which are constant here. Modified from (Nematzadeh, 2018).

We have shown here that the result of the detected houghlines matches the tilt cues as they appear in multiple scales of the DoG edge map of the pattern. This is exactly what we had aimed to find an explanation for the tilt effect in the pattern, highlighting different groupings of pattern elements at different scales of the edge map, as we can observe in the edge map representation.

The bottom of Figure 15, shows the detected houghlines for the Complex Bulge pattern in Green, displayed on the edge map at multiple scales. The EMap is shown for eight different scales from $\sigma_c$= 1 to $\sigma_c$= 4.5, with incremental steps of 0.5. The hough parameters are provided in the figure caption. As the detected houghlines in Green show in the figure, at fine scales-top row, the detection of slanted line segments on the edge map starts at scale 1.5 ($\sigma_c$= 1.5) with their integration resulting in a central bulge from the center outwards. This is getting more clearer at scale 2.0 ($\sigma_c$= 2.0), with the detected tilted lines around the center with an impression of bulge and then the detection of the tiles' edges out of the central region with no superimposed dots.

As the scale increases ($\sigma_c$= 3 to 4.5-second row), hough detects other line segments in nearly diagonal orientations (positive and negative), originating at the center whose integration results in an X shape grouping of these lines. Again our proposed hough analysis, correctly detects the tilt cues as they appear in the DoG edge map of this pattern at multiple scales. The result of this investigation on these illusory patterns has been delivered in a workshop poster[3].

## 6.3.    Systematically generated patterns

Similar to the Complex Bulge pattern, which has superimposed dots on a checkerboard, we designed and coded a set of simpler patterns generated systematically in MATLAB. Our focus was to minimize any additional tilt cues or secondary factors contributing to the illusory tilts seen in such patterns and stick to the main features, these include keeping the distribution of superimposed dots on the checkerboard nearly the same all over the pattern and leaving out different orientations for the configuration of the superimposed dots on the tiles. The secondary factors referred to, may have a high impact on the tilt effect perceived in many Tile/Tilt illusions. At the top of Figure 16 we have shown two patterns with these secondary tilt factors, 'Trampoline' and 'Straight dice' patterns. The tilt effect is quite strong in these patterns due to the changes of orientation and arrangement of the superimposed dots within the tiles. We have eliminated these factors and generated a set of systematic Tile illusions two of which are shown in the bottom row of Figure 16. We named these patterns 'Cross Bulge' and '1-3 dot' Tilt. These patterns have been investigated by our Vis-CRF model by extracting their local tilt cues from the EMap-DoGs in which reveals the underlying mechanism involved in the illusory tilt effect seen in these illusions. As noted before, Vis-CRF model is a generalized vision model and its output which is the edge map at multiple scales can be found for any type of input pattern/image.

[3] Nematzadeh, N., Powers, D. M., & Lewis, T. (2016) "A neurophysiological model for Geometric visual illusions".NeuroEng 2016: 9th Australasian Workshop on Neuro-Engineering and Computational Neuroscience.

For generating these patterns, we either repeated the same configuration designs for the superimposed dots on the same coloured tiles of a checkerboard (such as in Cross Bulge pattern), or applied a mirror symmetry/reflection in between even and odd rows (for instance in 1-3 dot Tilt pattern). In Figure 16 next to these patterns, we have also shown the enlarged cropped sections of these patterns for an easy navigation on these illusions. In these patterns the resolution of tiles are $100 \times 100$px and for dots it is $20 \times 20$px. There is a gap distance of 5px from the border of each tile to the location of dots. In the second order tilt illusions such as Complex Bulge illusion or Cross-Bulge pattern, the impression of the bulge effect is the result of foreground superimposed dots on top of a checkerboard (Nematzadeh, Powers and Lewis, 2017; Nematzadeh, Lewis and Powers, 2015).

In Cross Bulge pattern, each black square contains four white dots at its corners with the gap distance noted before (5px here). The impression of the illusory bulge or the expansion of white tiles in the pattern are quite strong and corners of black tiles appear to contract as a result of this illusion. In 1-3 dot Tilt pattern, the position of dots on the same coloured tiles have vertical symmetry from one column to the next, giving an impression of vertical waves along tile edges (connections of tiles' borders) which is classified as a strong tilt effect. Some convergence and divergence along the horizontal edges in this pattern are also observed.

We should note that different position of dots as well as their size result in various degrees of tilt effect (tilt, bow, wave; strong or weak) as well as expansion and contraction on checkers corners. We will explain how to adjust the range of scales ($\sigma_c$) in Vis-CRF model considering the sizes of image features for extracting the edge maps shortly, before the end of this section.

To highlight the effect of superimposed dots on top of a checkerboard, we have simply shown a small 3×3 symmetrical checkerboard in Figure 17. Based on the relative size of superimposed dots and the tiles of the checkerboard, we see the impression of central bulge very clearly for the given pattern. It is obvious that decreasing the size of superimposed dots results in less persistency of dot cues at fine scales of the DoG edge map, and a weaker illusory bulge effect at the center.

We illustrate the effects of a gap parameter which is the distance between the positions of dots to the tile borders (checkers) on top of Figure 17. As shown in the figure by decreasing the gap distance, we encounter with a weaker bulge effect. At the center of the figure we have shown the wave effect that still exist in two very small cropped versions of a tile illusion with superimposed dots on a checkerboard. Therefore the effect is due to the emergence of local tilt cues. The bottom of the figure shows the effect of rotation on the bulge effect which is invariant in respect to this factor. All of these parameters/features and similar scenarios should be tested for by conducting psychophysical experiments for a conclusive report on the tilt effect perceived in these illusions.

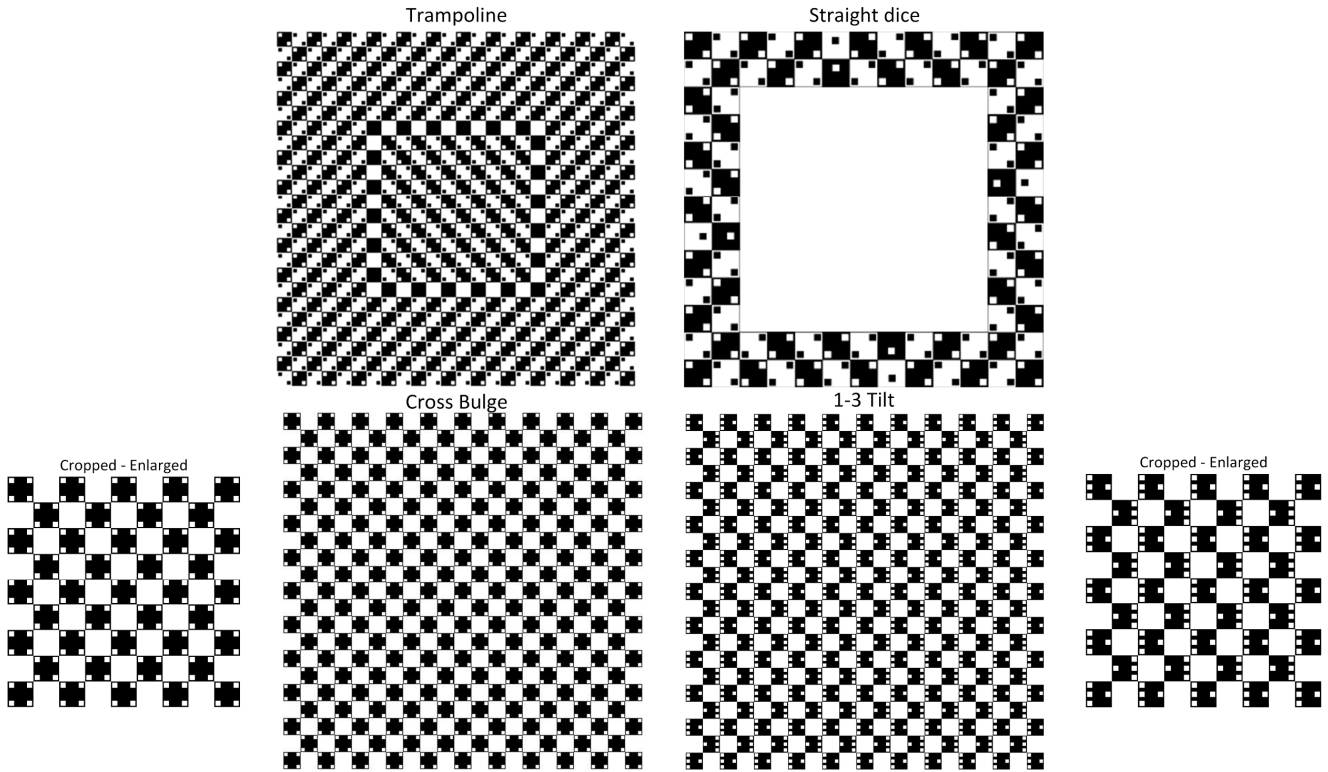

**Figure 16** Top: Trampoline (http://www.psy.ritsumei.ac.jp/~akitaoka/trampolineL.jpg, http://www.ritsumei.ac.jp/~akitaoka/motion-e.html) and Straight dice (http://brainden.com/line-illusions.htm#prettyPhoto[pp_gal]/13/) patterns with strong tilt effect. Bottom: Two samples of systematically generated Tile illusions by the lead author: Cross-Bulge and 1-3 dot Tilt patterns. The enlarged cropped versions of these patterns are provided for easy referral.

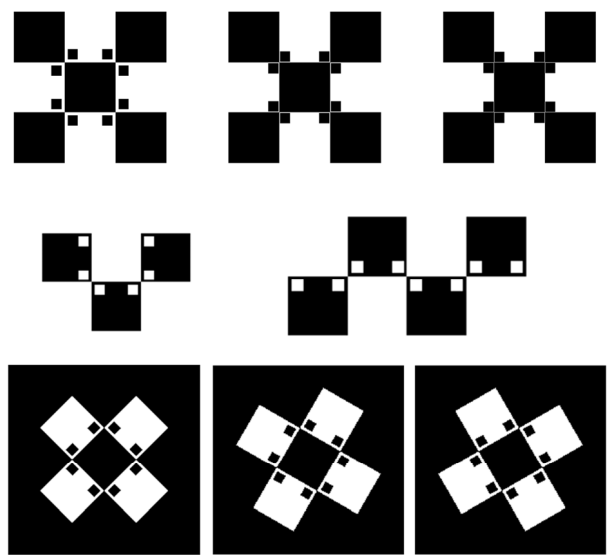

**Figure 17** Top: The effect of gap parameter on the central bulge effect on a 3×3 checkerboard. Center: Wave tilt effect in two cropped versions of a Tile illusion with superimposed dots on top of a checkerboard. Bottom: Rotation invariance for the central bulge effect on a 3×3 checkerboard.

Figures 18 and 19 show the edge maps of Cross Bulge and 1-3 dot Tilt patterns with the size of 9×9 tiles (Tiles of 100×100px and dots are 20×20px with the gap distance of 5px from the border lines of each tile). So the smallest feature we need to detect are the dots and their borders as well as sharp edges around the tiles. The following calculations is done to find the "range of scales" for generating the edge maps based on the center and surround Gaussians. This way we considered around 95.6% of the center and exactly 68.3% of the surround Gaussian in the edge map representation ($s$=1.6, h=8).

Dots: $2 \times \sigma_s = 20 \rightarrow 2 \times s\, \sigma_c = 20\text{px} \rightarrow \sigma_c = 20/(2 \times 1.6) \rightarrow \sigma_c = 6.25$      (rounded to 5)

Tiles: $2 \times \sigma_s = 100 \rightarrow 2 \times s\, \sigma_c = 100\text{px} \rightarrow \sigma_c = 100/(2 \times 1.6) \rightarrow \sigma_c = 31.25$      (rounded to 30)

For sharp edges: $2 \times s\, \sigma_c = 1\text{px} \rightarrow \sigma_c = 1/(2 \times 1.6) \rightarrow \sigma_c = 0.3125$      (rounded to 0.5)

So the range of scales would be: 0.5 to 30 with incremental steps of 0.5. To reduce the size of the edge map (# of layers) we can consider the range of 1 to 30 with incremental steps of 1, which reduces the size of EMap-DoGs to half. Perceptual grouping under this range also reveals the information we are looking for (The only difference in encoding is that the model now can detect edges with the thickness larger than 2px).

In Figures 18 and 19, we have shown the DoG filters as well as the edge maps at five scales $\sigma_c$ = 1, 5, 10, 15, 20 for these two Tile illusions (Cross Bulge and 1-3 dot Tilt patterns). The scales that highlight the illusion and the local tilt cues contributing to the illusory tilts are scales 5 and 10 ($\sigma_c$ = 5, 10) where the center Gaussian can capture superimposed dots in the edge map representation.

The "*perceptual groupings*" revealed in the edge maps across multiple scales, easily cover the whole range of visual representation from fine details of the edges shown for scale 1, and then at the coarse scales such as scales 20 and 25 ($\sigma_c$ = 20, 25), we can get the same impression of groupings of pattern features when **squinting**, resulting in a *blurred version* of these patterns. These groupings can be revealed also by *increasing our distance* from the patterns that results in reducing their visual angles. This illustration highlights the efficiency and accuracy of the Vis-CRF model in encoding the visual input and showing the reason behind the use of a multiple scale edge map representation and not one multiscale output for a complete vision model.

Figure 20 shows the EMap-DoG of a pattern called 'Splendid spiral' (Flicker, Website). This pattern is not a Tile illusion but we included this here in our research, since the edge map and perceptual grouping of the dots across multiple scales follows an interesting pattern and the EMap-DoG can address the illusion we can observe. Variations of this illusion can be found based on checkerboard design, or polygonal dots at Pinterest website[4]. In this illusion, dots are arranged

---

in a circular formation with a horizontal translation that impact more heavily on the position of inner circles compared to the outer ones. The direction of shift keeps changing every few steps and due to this we encounter with a strong spiral effect rather than any circular arrangements of the dots. The varying distance between dots in each circle also contribute to the spiral perception we see in this illusion. The impression of a 3D wave-like surface is also observed in this pattern.

For generating the edge map, we have selected the range of scales from $0.5$ to $5.0$ with incremental steps of $0.5$, considering the pattern features ($\sigma_c$ = 0.5, 1.0, 1.5, ..., 4.5, 5.0). Dot sizes are in the range of $11px$ to subpixel in the center. In Figure 20 we have shown the EMap-DoG in two columns, in which the first column on the left shows fine to medium scale DoGs and the second column (the one on the right) shows the coarse scales. As illustrated in the edge map, the center of the spiral started to emerge at medium scales, and across these scales, the EMap shows the illusory wave having 3D motion. As we move to more coarse scales in the edge map, the spirals gets bigger, extending from the center to the outer region of the pattern, therefore we see the spiral at our global view and by focusing to the pattern we encounter with $3D$ wave motion, upwards and downwards. This example illustrates that simple DoG filtering, with precise parameters that seen in Vis-CRF model, is the underlying mechanism for encoding some of the depth information as well as the contrast and illumination within this intermediate representation, as Marr called this encoding a primal sketch having $2.5D$ information. Our model implements the activation responses of simple cells and their lateral inhibition effect and we explain more about Marr's edge detection in Section 7.

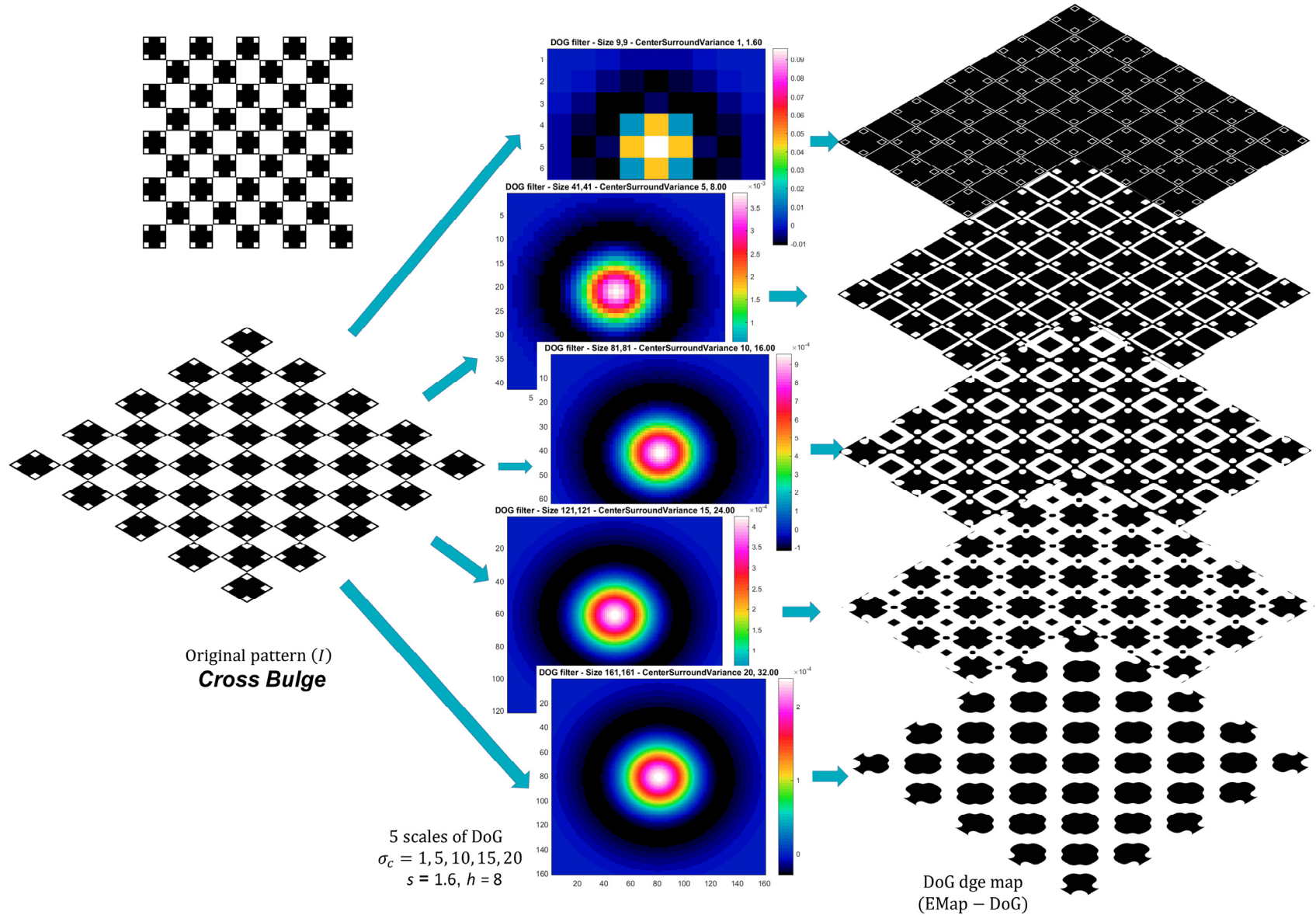

Original pattern (*I*)
***Cross Bulge***

5 scales of DoG
$\sigma_c = 1, 5, 10, 15, 20$
$s = 1.6, h = 8$

DoG dge map
(EMap − DoG)

**Figure 18** EMap-DoG of a crop section of 'Cross Bulge' pattern with the size of 9×9 tiles, (tiles of 100×100px, square dotes of 20×20px and the gap of 5px from tiles' borders) (Nematzadeh, 2016 ©). The scales of DoG filters are $\sigma_c$= 1, 5, 10, 15, 20. The fundamental parameters of the model are $s$=1.6, and $h$=8 (Surround and Window ratios respectively).

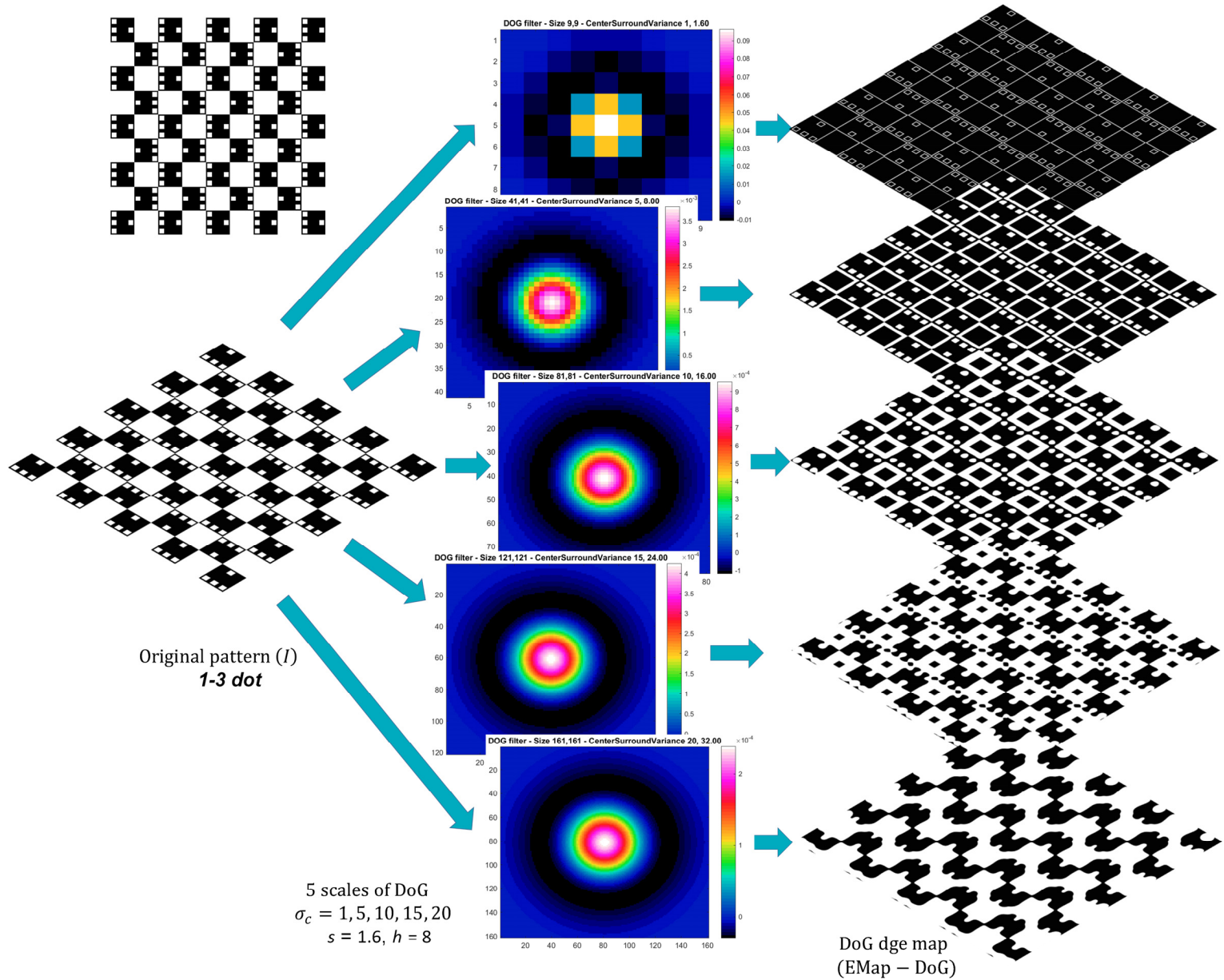

**Figure 19** EMap-DoG of a crop section of '1-3 dot Tilt' pattern with the size of 9×9 tiles, (tiles of 100×100px, square dotes of 20×20px and the gap of 5px from tiles' borders) (Nematzadeh, 2016 ©). The scales of DoG filters are $\sigma_c$= 1, 5, 10, 15, 20. The fundamental parameters of the model are $s$=1.6, and $h$=8 (Surround and Window ratios respectively).

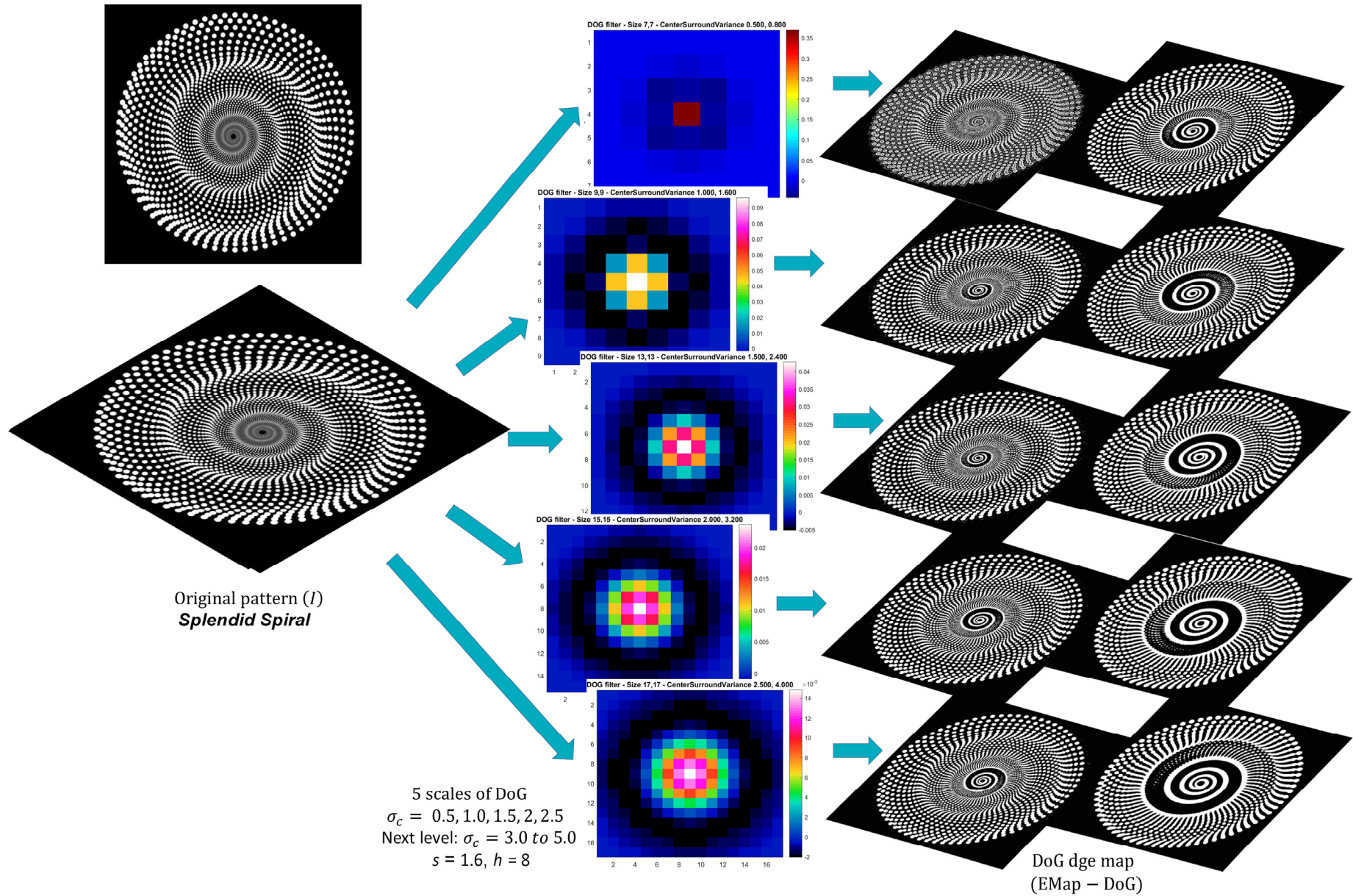

Original pattern (*I*)
***Splendid Spiral***

5 scales of DoG
$\sigma_c = 0.5, 1.0, 1.5, 2, 2.5$
Next level: $\sigma_c = 3.0\ to\ 5.0$
$s = 1.6, h = 8$

DoG dge map
(EMap − DoG)

**Figure 20** EMap-DoG of 'Splendid Spiral' pattern (https://www.pinterest.com.au/pin/402298179198480342/visual-search/) with the size of 500×500px. The size of dots are in the range of 11px to subpixel in the center. The scales of DoG filters are $\sigma_c$= 0.5, 1.0, 1.5, 2.0, ..., 4.5, 5.0). The fundamental parameters of the model are $s$=1.6, and $h$=8 (Surround and Window ratios respectively).

# 7. Extension of the DoG model to LoG level

## 7.1. Marr's theory of edge detection

Hubel and Wiesel's (1962) description of simple cells was explained by the bar- or edge-shaped receptive fields. This led to a view of the population of feature detectors of edges and bars of various widths and orientations in the visual cortex explained by Barlow (Barlow, 1969). Another view was from Campbell and Robson's (Campbell and Robson, 1968) experiments that presented images in parallel with a number of independent orientation and spatial frequency-tuned channels. This view considered the visual cortex as a kind of spatial Fourier analysis. Based on these two perspectives, It was noted by Marr and Hildreth (1980) that none of these approaches can provide any direct information about the purpose of the early analysis of an image.

Marr and Hildreth emphasized that "the purpose of early visual processing is to *construct a primitive but rich description of the image* that is to be used to determine the reflectance and illumination of the visible surfaces and their orientation and distance relative to the viewer" (Marr and Hildreth, 1980- p.188). The first primitive descriptor stated by Marr in his theory of edge detection is called the "Primal sketch" (Marr, 1976) which consists of two subcategories. First, the intensity changes that defines edge segments, bar, blob and termination. Second, the geometric relations with larger more abstract tokens, which are constructed by selecting, grouping, and summarising of the raw primitives in different ways. These two results in a hierarchy of descriptors that covers a range of scales, referred to as "full primal sketch". The primal sketch results in the formation of a satisfactory computational theory by Marr and his colleagues.

As specified in edge theory of Marr and Hildreth (1980), intensity changes arise from either one of the follows: surface discontinuities, reflectance or illumination boundaries and all of these properties are spatially localized. Furthermore, it can be the result of changes in the orientation and distance of observer from the visible surface. Wherever an intensity change occurs in an image there will be either a corresponding peak in the first directional derivative or a zero crossing in the second derivative of intensity (Marr, 1976b, Marr and Poggio, 1979).

The idea of multiscale edge analysis by Marr and Hildreth (1980) for producing edge maps of the image at different scales is well known in computer vision (CV) society. In their proposal, they emphasised that the optimal smoothing filter for images should be localized in both spatial and frequency domains. The localization requirements in spatial and frequency domain are conflicting: $\Delta x \Delta \omega \geq \pi/4$ due to the uncertainty principle (Bracewell 1965, p.160-163). There is only one distribution that optimizes this relationship which is the Gaussian (utilized in Vis-CRF model). Therefore, Gaussian filter is the only operator that satisfies the uncertainty principle and has the best trade-off for simultaneous localization of both spatial and frequency domains. Marr and Hildreth have suggested that by applying Gaussian filters at different scales to an image, we can obtain a set of images with different levels of smoothness referred to as *edge maps* (Basu, 2002; Marr and Hildreth, 1980). In order to detect the edges in the edge map they used zero-crossings of the second derivatives of Gaussian which is the Laplacian of a Gaussian (LoG) function.

The processing of natural images are covered in this section and have their own complexities since the studied intensity changes cover a wide range of scales. The intensity changes best detected by finding the zero crossings of Laplacian of Gaussian for an image (second derivative of Gaussian; $\nabla^2 G(x, y) * I(x, y)$ where $\nabla^2$ is Laplacian and $G$ is the 2D Gaussian) (Marr & Hildreth, 1980). For scenery images there is no single filter that can be optimal simultaneously at all scales since the changes occur at different scales. Marr and Hildreth (1980) noted that for this we should first take local averages at various resolutions and then detect the changes in intensity that occur at each one. We need also to reduce the range of scale and as recommended by Marr and Hildreth (1980), the filter spectrum should be smooth and roughly band limited in the frequency domain (its variance, $\Delta\omega$ should be small) (Marr and Hildreth, 1980). The second derivative of Gaussian filter (LoG) does not need to be orientation-dependent. The intensity changes in each of the channels are then presented by zero-crossing segments. Marr and Hildreth (1980) further noted that the zero crossing segments from different channels are not independent and in the image description they should combine based on predefined rules. This create an oriented primitive descriptor. This description noted to be a 'raw primal sketch'.

Marr and Hildreth noted that: "To detect changes at all scales, it is necessary only to add other channels, like the one described above, and to carry out the same computation in each. These representations are precursers of the descriptive primitives in the raw primal sketch, and mark the transition from the analytic to symbolic analysis of an image. The remaining step is to combine the zero crossings from the different channels into primitive edge elements." (Marr and Hildreth, 1980 - p. 195)

There is a constraint for the contributions of each point in the filtered image. It should arise from a smooth average of nearby points rather than any kind of average of widely scattered points (it should be smooth and localized in the spatial domain), since "The visual world is not constructed of ripply, wave-like primitives that extend and add together over an area" (Marr 1970 p. 169). It is actually constructed from contours, creasers, marks, scratches, shadows, and shading (Marr and Hildreth, 1980- p. 189).

Marr and Hildreth in their theory of edge detection showed that forming oriented zero crossing is possible from the output of center surround Laplacian filters ($\nabla^2 G$) on the image. We have also shown similar output from the center surround DoG in Vis-CRF model when illusory tilts in some Tile/Tilt illusions were previously explained (refer to p. 21-35 ) (Nematzadeh and Powers, 2019;

Nematzadeh, 2018; Nematzadeh, Powers and Lewis, 2017). This is the basis for a physiological model of simple cells (Marr and Ullman, 1979).  Marr's edge detection provides a raw primal sketch. We want to emphasise here that Vis-CRF can automatically detect perceptual groupings, (Nematzadeh, Powers and Lewis, 2017) and to some extent the geometrical relations in the intermediate representation of the edge map alongside the intensity encodings in studied images. As noted by Marr, this should be done by summarising the raw primitives in the edge map and we have shown this across multiple scales of EMap-DoGs (and also EMap-LoG; that will be discussed later).Therefore, the edge map representation in our Gaussian RF model is a rich descriptor as defined by Marr's and Hildreth's (1980) theory of edge detection and the primal sketch description in their theory.

The aggregate response of a group of ON- and OFF- receptive fields can produce the directional selectivity in the response as demonstrated by Marr and Hildreth  (1980): "If P presents an ON-center Geniculate X-cell receptive field, and Q, an OFF-center one, then if both are active, a zero-crossing Z is the Laplacian passing between them. If they are connected to a logical AND gate, as shown, then the gate will 'detect' the presence of the zero-crossing. If several are arranged in tandem, as in (b) and also connected by logical ANDs, the resulting operation detects an oriented zero-crossing segment within the orientation bounds given roughly by the dotted lines. This gives our most primitive model for simple cells. Ideally one would like gates such that there is a response only if all (P, Q) inputs are active, and the magnitude of the response then varies with their sum" (Marr and Hildreth, 1980-Fig. 9 Caption).

### 7.2. Vis-CRF Characteristics

We hypothesize that visual perception of a scene starts by extracting the multiscale edge map, and proposed that a bioplausible model of *contrast sensitivity* of retinal GCs using DoG filtering produces a stack of multiscale outputs (Romeny, 2008) as an intermediate representation. The contrast sensitivity of the retinal ganglion cells can be implemented based on *Classical Receptive Field* (CRF) *models*, involving circular center and surround antagonism. To extract and reveal the edge information it uses the differences and second differences of Gaussians (Enroth-Cugell and Robson, 1966; Rodieck and Stone, 1965) or Laplacian of Gaussian (LoG) (Ghosh, Sarkar and Bhaumik, 2007). As noted before Marr and Hildreth (1980) proposed an approximation of LoG with DoG, based on a specified ratio of standard deviations ($\sigma$) of the center and surround Gaussians. Symmetrical DoGs as well as normalized ones at multiple scales have been used for spatial filtering in our study (with the area under the curve being zero) to generate the edge maps (EMap-DoGs), for modelling ON-center OFF-surround receptive fields (RFs). It has been shown that Vis-CRF model can predict the illusory tilt in the family of Café Wall patterns by providing both qualitative and quantitative results as well as for a set of Tile illusions (Nematzadeh, 2018; Nematzadeh and Powers, 2019; Nematzadeh and Powers, 2017c).

Vis-CRF model has resemblance to Robson's (1983) model which found a retinal ganglion cell (GC) response to the pattern (image). He demonstrated that a model of ganglion cells with the antagonistic center and surround organization based on DoG can predict the cell response to *any stimulus* pattern. The convolution between the pattern's luminance function with the differences of Gaussians should be found for this as the weighting function of ganglion cells receptive fields (RFs). On the other hand the multiple scale analysis in Vis-CRF model is inspired by the proposed foveal retinal vision by Lindeberg and Florack (1994). Their model is based on the simultaneous sampling of the image at all scales, and the edge map at multiple scales (EMap-DoG) in our model is generated in a similar way.

Vis-CRF model is the simplest simulation of all for the classical receptive fields (retinal/cortical), using isotropic differences of Gaussians (DoG). In comparison to Marr's and Hildreth's edge descriptor based on finding the zero-crossing segments of the LoG filtered output, we have implemented a DoG approximation to it. This is done with a specified scale ratio of the center and surround Gaussians as described by them ($\sigma_i/\sigma_e$ = 1.6 (Marr and Hildreth, 1980); where i is inhibitory/surround, and e is excitatory/center Gaussians). This is the fundamental parameter of Vis-CRF model and the binary edge map in our simulations is quite similar to the LoG output of Marr's and Hildreth's model prior to finding the zero-crossings. Therefore the unique characteristic of the model is that it approximates the Laplacian of Gaussian (LoG) by its specific parameter ($s=1.6$) without adding any computational complexities to the model.

On the other hand, setting up the model parameters or fine tuning it is easy to handle. For instance, we set up the range of scales empirically so far by manually checking image features and their resolutions considering the definition of DoG kernel and the model parameters ($s=1.6$, h=8). Note that *scale invariant processing* in general is not sensitive to the exact parameter setting; ideally the parameters of Vis-CRF model ($\sigma_c$, $s$, $h$, range of scales) should be set in a way that, at fine scales, the edge map can capture high frequency details and at coarse scales, the kernel should have appropriate size relative to the objects of interest within the scene. This can be automated if we set up a threshold for optimizing the DoG response locally after applying the normalization phase at each scale. This process is called scale adjustment based on the image/pattern features for setting up the model parameters.

Displaying high precision details in the EMap-DoGs of our simplified Gaussian RF model is the key success. We have shown edge map representations of Vis-CRF model for a range of Geometrical illusions so far both in this work and our previous research (Nematzadeh and Powers, 2019; Nematzadeh, Powers and Lewis, 2017; Nematzadeh and Powers, 2017c; Nematzadeh and Powers, 2016b) and the model performance on a scenery image as our general vision task later in this section. We also tested our model on a range of Medical images (MRIs and scans) as well as a few samples of other imaging types such as thermal and infrared (IR) imagery and we found very promising results that guaranteed the generalization of the model for a diverse range of applications. Some image enhancement techniques may be required to increase the precision of the detection in our analysis (based on the application) such as for Medical images. We will concentrate on these other applications in our future research and will publish some of the preliminary investigations we had done so far in our upcoming publications soon. To summarise, we want to note that Vis-CRF is a reliable model for encoding a rich descriptor for an image in general at low-level stages in any application that requires high clarity of the edge points. The edge map at multiple scales can then be used to determine the most suitable processing stages at their latter processing (higher levels) specified by the application.

Like Marr's edge detection model, we should note that Vis-CRF model is not a complete proposal for a physiological mechanism of vision since it ignores the attribute of directional selectivity of more complex cells – It is just a unique simulation of retinal/cortical simple cells with its fundamental parameter of $s = 1.6$, considering the intensity changes only, and it achieves accurate results in perceptual grouping as well. In our analysis we have not implemented different channels for colour in the model yet, therefore it is not a full descriptor.

Let's move to our investigation on a scenery image. Figure 21 shows the DoG edge map of Vis-CRF model for a sample natural image shown on the top-left of the figure. We labelled the roses in the image from 1 to 4 for easy referral in our explanations in the text. The image size is $2250 \times 1644$px, and the fundamental parameters of the model are $s = 1.6$, h = 8 for the Surround and Window ratios respectively. The range of scales for the edge map are from $0.25$ to $10$ with incremental steps of $0.25$ for a complete analysis of the edge map in the image. Though a limited number of scales are shown in the figure due to shortage of space ($\sigma_c = 1$ to 9 with incremental steps of $1$). The range of the DoG scales is selected empirically in this experiment in a way to capture nearly a full range of fine to medium scale features from the image. To extract an optimized and rich edge descriptor, we can use Logan's theorem (1977) as described by Marr and Ullman (Marr, Ullman and Poggio, 1979) to find the zero crossing of one-octave band pass signals in which the set of such zero crossing segments would be *extremely rich in information* (Marr and Hildreth, 1980). The bottom left corner of Figure 21 shows the enlarged cropped sections from Rose#3 at four scales ($\sigma_c = 1$ to 4 with incremental steps of $1$). As indicated in the figure, Vis-CRF model can detect the edges of dew on top of the flower petal, although this section of the image is quite blurry. It can also reveal the petal texture of Rose#1 clearly at medium scales of the edge map. This indicates high precision edge map representation in our DoG based model.

Figure 22 shows the DoG edge map in Jetwhite colormap (Powers, 2016) presented at twenty different scales here ($\sigma_c = 0.5$ to $10.0$ with incremental steps of $0.5$) for the sample natural image given on the top left. An enlarged version of the DoG at scale seven ($\sigma_c = 7.0$) has shown for this pattern on the top right next to the image. The colour bar next to the edge map facilitate navigation around the edges of the image and provides further detail that will allow the reader to develop additional insights to what is happening here. The enlarged versions of two cropped samples from Rose#1 and Rose #3 are presented at the top, facilitating a comparison between a blurry edge in Rose#3 and sharp edges in Rose#1 at one scale of the DoG edge map ($\sigma_c = 7.0$) in the jetwhite colormap. Vis-CRF can detect the blurry edges of Rose#3 petal as well as fine details such as petal structure of Rose#1 at medium scales, although sharp edges are distinguished easier at fine scales.

For the edges encoded in the DoG edge map representation, we have positive and negative components next to each other; due to the shape of Gaussian kernel. This has some resemblance to the zero crossing in Marr's theory of edge detection using the LoG edge map.

### 7.3. Extension of Vis-CRF model to LoG level

As noted before, wherever an intensity change occurs in an image there will be either a corresponding peak in the first directional derivative or a zero crossing in the second derivative of intensity (Marr, 1976; Marr and Poggio, 1979). For this Marr and his colleagues found zero crossing of the second derivative $\nabla^2$ of intensity in the appropriate direction.

$G''$ looks like a Mexican hat operator, closely resembling Wilson and Giese's (1977) differences of two Gaussians (DoG) that is used to generate the LoG edge map representation at one level higher than EMap-DoG in our model. In fact, this technique is the same as finding the limit of the DoG function as the sizes of two Gaussians tend to one another. Since a gradual change of scales is used in our model, a reliable result is calculated by this approximation. It is noted that LoG approximately generates a bandpass operator with a half power bandwidth of about 1.2 Octaves (Marr and Hildreth, 1980).

Therefore, in our analysis on the sample scenery image, we tested also this extension of the model to find out the differences of EMap-DoGs at different scales in order to generate the edge map at LoG level for further analysis of the edges in the image. We simply need to find the differences between the scaled EMap-DoGs at each consecutive scales from the DoG edge map to generate EMap-LoG. The second derivative of Gaussian (LoG) is given in (4) and as noted before, it can be estimated by the Differences of two DoGs from eq. 1 as well.

$$\nabla^2 G(x,y) = d^2/dx^2\, G(x,y) + d^2/dy^2\, G(x,y) = (x^2+y^2-2\sigma^2)/\, 2\pi\sigma^6\, exp[-(x^2+y^2)/(2\sigma^2)] \qquad (4)$$

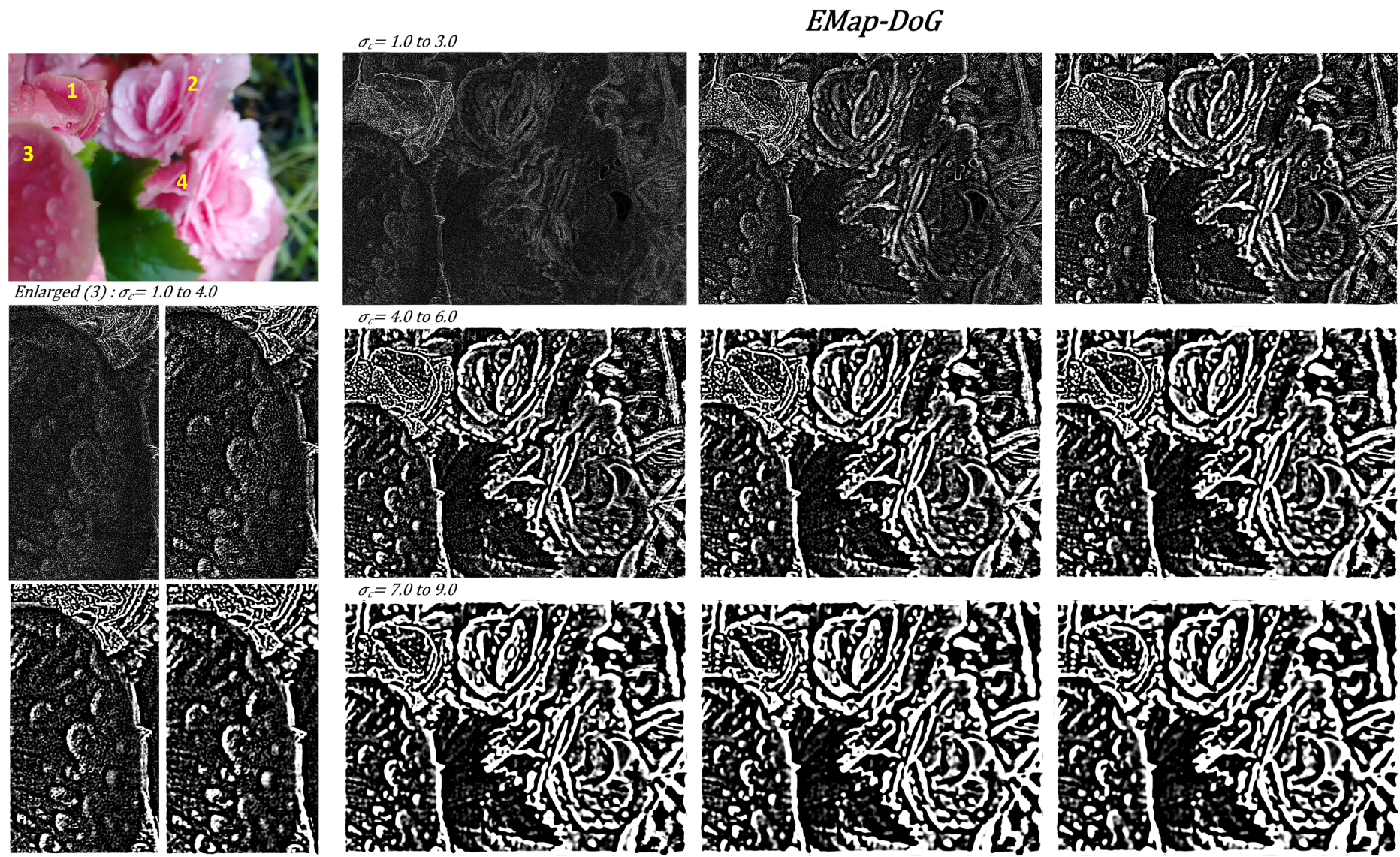

*EMap-DoG*

**Figure 21** The DoG edge map presented at nine different scales ($\sigma_c = 1$ to 9 with incremental steps of 1) for a sample of natural image on the top left. The bottom left corner shows the enlarged cropped sections from Rose#3 at the first four scales on the right ($\sigma_c = 1$ to 4). Vis-CRF model can detect the edges of dew on top of the flower petal as well as the petal's texture clearly at medium scales although the image at this part is quite blurry. The image size is 2250×1644px, and the fundamental parameters of the model are $s = 1.6$, $h = 8$ (Surround and Window ratios respectively). We should note that for a complete analysis of the edge map we have tested a range of $\sigma_c = 0.25$ to 10.0 for the scales, and due to the shortage of space we have shown a limited number of scales in the figure.

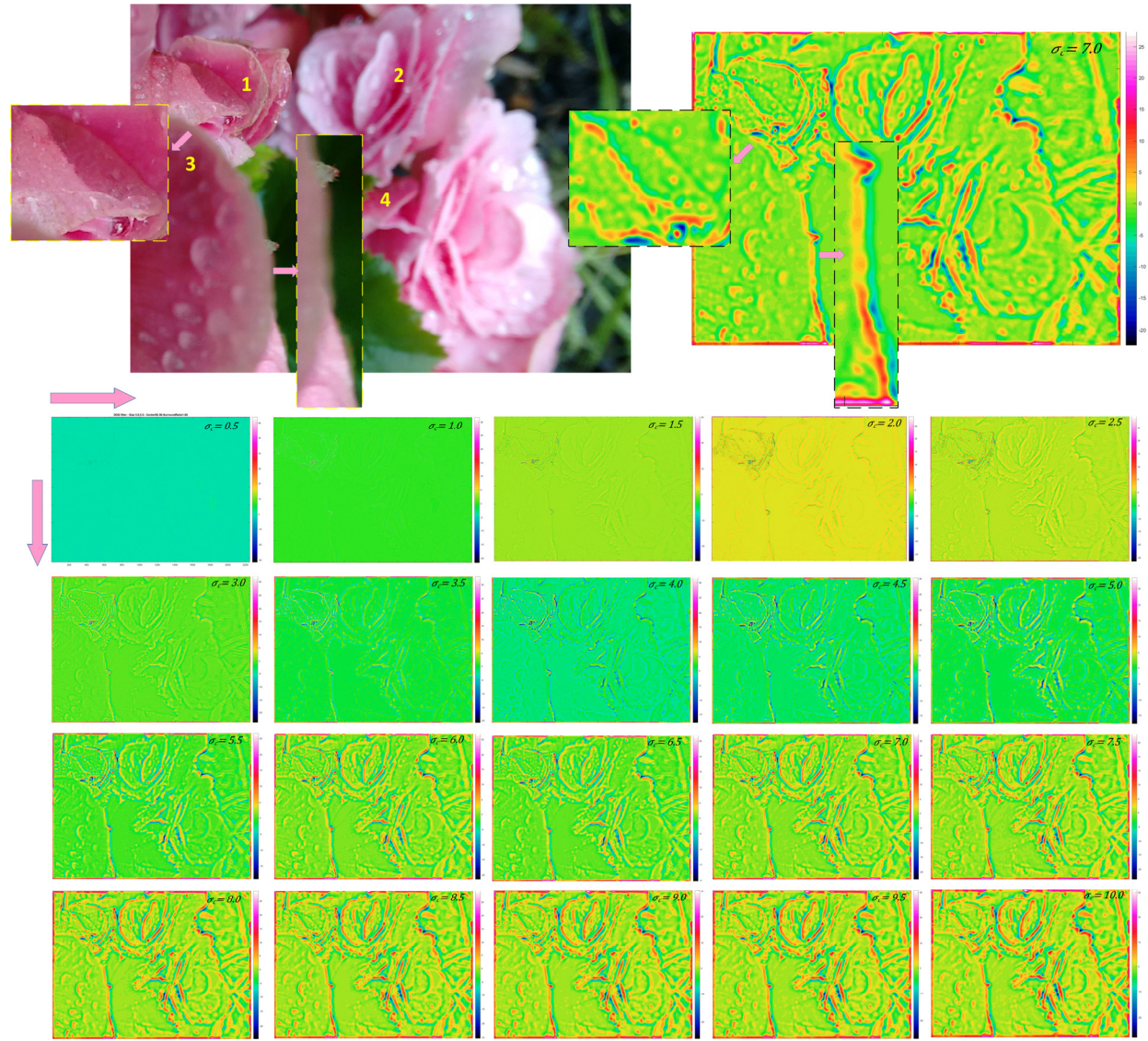

**Figure 22** The DoG edge map in Jetwhite colormap (Powers, 2016) is presented at twenty different scales here ($\sigma_c = 0.5$ to $10.0$ with incremental steps of 0.5) for the sample natural image given on the top left. An enlarged version of the DoG at scale seven ($\sigma_c = 7.0$) for this pattern has shown on the top right next to the image. The enlarged versions of two cropped samples from Rose#1 and Rose #3 are provided at the top, letting us to compare the difference between a blurry edge in Rose#3 and sharp edges in Rose#1 at one scale of the DoG edge map. Vis-CRF model can detect the blurry edges of Rose#3 petal as well as Rose#1 petal structure at medium scales. The sharp edges will be clearer at fine scales. In the DoG edge map for the edges, we have positive and negative components next to each other; due to the shape of Gaussian kernel. This is the zero crossing as Marr defined in his edge theory based on the LoG edge map. The fundamental parameters of Vis-CRF model are $s = 1.6$, $h = 8$ (Surround and Window ratios respectively).

The Laplacian ($\nabla^2$) is the only orientation-independent and second-order differential operator. If we use Logan's theorem (Logan Jr, 1977) for zero-crossings of one-octave bandpass signals, the set of such zero crossing segments is extremely rich in information as said before. If the filters had bandwidth of one octave or less as it was proved in the Logan's theorem, they would in fact contain complete information about the filtered image (Marr and Hildreth, 1980). $\nabla^2 G$ filter has a half sensitivity bandwidth of about 1.75 octaves, which puts it outside the range in which Logan's theorem applies as noted in (Marr and Hildreth, 1980).

Two types/layers of edge map in our Vis-CRF model are shown in Figure 23; EMap-DoG is presented in the middle and EMap-LoG on the right. The range of scales are from $0.25$ to $10$ with incremental steps of $0.25$ for a complete analysis of the image edge map. We then selected a limited number of scales from the edge map to be shown in the figures due to shortage of space (here $0.5$ to $4.5$ with incremental steps of $0.5$). Rather than using the LoG formula given in (4) (with higher complexity), LoG was calculated at the second layer by finding the differences between two consecutive scales in the DoG edge map. This is feasible since for encoding the visual input in the edge map, the step sizes are set up with gradual changes in the scales of the DoGs. Vis-CRF model highlights the most dominant edges in the LoG representation. Two enlarged versions for a particular scale of DoG ($\sigma_c = 4.0$) and one level of LoG [DoG ($\sigma_c = 4.0$) - DoG ($\sigma_c = 3.75$)] are shown next to the image on the left side of the figure.

To explore more on the LoG response, we have shown the LoG edge map (EMap-LoG) in Figure 24, presented at nine different scales (the differences in between DoG scales are from $\sigma_c = 1$ to $9$) for the sample scenery image on the top left. We considered the whole range of scales which has been used for generating the DoG edge map ranging from $0.25$ to $10.0$ with incremental steps of $0.25$ for displaying the LoG edge map in the figure. For instance at the first level of LoG, we have calculated [DoG ($\sigma_c = 1.25$) - DoG ($\sigma_c = 1.00$)] not [DoG ($\sigma_c = 2.00$) - DoG ($\sigma_c = 1.00$)]. We have done this to highlight the effect of gradual changes of scales with the step size of $0.25$ in our model (we emphasize again that the reduction of the number of scales in the figures are just for displaying purposes only). The bottom left corner of the figure shows the zoomed in versions from Rose#3 and Rose#1 at the first four scales in the figure. Vis-CRF model can detect *'dominant edges'* [5] in the pattern where the edges are clear in the LoG edge map or in the DoG edge map across a few consecutive fine scales. We can see the sharp edges in Rose#1 at fine scales as well as some edges that revealed for dew on top of Rose#3 although the image at this part is quite blurry, but due to slight intensity changes with the flower petal, the model can detect these (The image size is $2250 \times 1644$px, and the fundamental parameters of the model are $s = 1.6$, h = 8, Surround and Window ratios respectively).

---

[5] If an edge appears across multiple fine scales of the DoG and/or LoG edge maps, we call it a dominant edge.

To display the LoG edge map with more details, we have shown these in the Jetwhite colormap (Powers, 2016) as presented at nineteen different scales in Figure 25. The range of scales is from $0.25$ to $10.0$ with incremental steps of $0.25$ in our complete analysis of the image to generate the edge map but similar to previous figures, we have selected a limited range of scales to be shown in the figure for the sample natural image given on the top left. As noted before, the LoG edge map is calculated by finding the differences between consecutive scales in the EMap-DoG. An enlarged version of the LoG at level $\sigma_c = 7.0$ [DoG($\sigma_c = 7.25$) - DoG($\sigma_c = 7.00$)] has shown for this pattern on the top right next to the image. The zoomed in versions of two cropped samples from Rose#1 and Rose#3 are given at the top, to find out the difference between a blurry edge in Rose#3 and sharp edges in Rose#1 at one level of the LoG edge map. Vis-CRF can detect the blurry edges of Rose#3 as well as fine details such as Rose#1 petal structure at fine to medium scales.

An important attribute of the *edges* in the proposed DoG and/or LoG based edge maps is that it contains both positive and negative values due to the convolution with the DoG filter. We have not done any truncation of the negative values, nor any type of morphological operations so far on the output. For further analysing of the edges, the only thing that we may need at low-level stage is *thresholding of the results*. For instance defining a threshold out of the mean of positive (Thr-Max) and the mean of negative (Thr-Min) values to discard some of the non-important responses in the edge map, which is essential for an efficient and effective processing. We need thresholding for the reduction of the data processing or getting rid of the non-informative data, while increasing the speed of processing which is a key attribute in any bioplausible approach especially in modelling vision. There are some neuronal correlates and biological evidence to the thresholding. For instance Bekesy (1967) in his book of 'Sensory inhibition' noted that: "The simplest way to get rid of information is to reduce the sensitivity of the receptors. This method is used effectively in all complex living systems" (p. 9). He then continues by stating that "the process of masking and adaptation reduce information by reducing receptor sensitivity and this reduction continues even after the disturbing stimulus has ceased" (p.19). He then added that "a more suitable method of eliminating unwanted information is through the process of compensation" (Bekesy Von, 1967; p. 19). We leave further argument and explanations about finding the dominant edges from the edge maps in our model based on the mean thresholding of the positive and negative values (Maxes and Mins) in the edge map to another work since it is out of the scope of this research.

## Summary

The current models for Geometrical illusions are quite complicated and hence we need less sophisticated and more bioplausible approaches in detecting visual cues and clues for addressing these illusions. We believe that further exploration of the role of simple Gaussian receptive field (RF) models (Blakeslee and McCourt, 1999; Ghosh, Sarkar and Bhaumik, 2006; Marr and Hildreth, 1980) in front-end vision and low-level retinal processing will lead to more accurate computer vision techniques and models with the property of detecting features that humans see and detect. These effects can contribute later to higher-level models of depth and motion processing and generalized to computer understanding of natural images.

In this research, we further explored the neurophysiological model of low-level filtering at multiple scales developed by (Nematzadeh, 2018; Nematzadeh, Lewis and Powers, 2015), based on the circular center and surround mechanism of the retinal receptive fields. To filter, a set of the Differences of Gaussians (DoG) at multiple scales is used to model the multiscale Retinal Ganglion Cells (RGCs) responses to any arbitrary stimulus. The simulation output is an edge map representation at multiple scales for the visual scene/pattern, which is utilized to highlight the tilt effects in Tile/Tilt illusion patterns as well as visual cues and clues in natural images investigated in this research.

We explained here the formal descriptions and parameters of our Gaussian RF model (Vis-CRF), with more focus on the edge map representations at multiple scales as the output of the Vis-CRF model for the patterns/images being investigated. Further analysis[6] of the detected tilt cues in the edge map and quantitative measurement of the degree of tilt of these line segments can be found in (Nematzadeh, 2018).

Our results suggest that this multilevel filtering explanation, which is a simplified simulation for the Retinal Ganglion Cell's responses to these patterns is indeed the underlying mechanism connecting low-level filtering to mid- and high-level explanations such as the *Anchoring Theory* (Economou, Zdravkovic and Gilchrist, 2007; Bressan, 2006; Gilchrist, Kossyfidis, Bonato et al., 1999) and *Perceptual Grouping* (Nematzadeh and Powers, 2017b). The multiple scale edge map representation in our model has some analogy to Marr's and Hildreth's (1980) suggestion of retinal 'signatures' of the three-dimensional structure from a raw-primal sketch, this being supported by physiological evidence (Field and Chichilnisky, 2007; Martinez-Conde, Macknik and Hubel, 2004; Shapley and Perry, 1986).

Marr and Hildreth described how directional selectivity is possible through the aggregate response of a group of nearby ON- and OFF- receptive fields. The edge map representation in our simulations (EMap-DoG and LoG) is consistent with Marr's theory of edge detection and his speculation of the simple cells encoding to include directional selectivity properties in the edge map. We have shown this for the illusory tilts we perceive in Tile/Tilt Illusions in general and some illusion effects in other Geometric Illusions in particular such as the Spiral Café Wall and Splendid Spiral described here and for Hermann Grid and Zöllner illusion, explained in (Nematzadeh, Powers and Lewis, 2019). Based on the experimental results, we want to emphasize that more sophisticated models of non-classical receptive fields (nCRFs) based on elongated/anisotropic filters such as (Todorovic, 2017) are not essential to reveal the illusory tilts in these patterns. We should note that for the final perception of these illusory tilts and the integration of the local tilt cues, these types of models are useful for higher level cortical processing by more complex cells. We also showed that in all these illusions investigated, two or more incompatible groupings of pattern structure arise simultaneously in their edge maps as a result of our local and peripheral views of these stimuli (Nematzadeh, Powers and Lewis, 2017). That is a major factor for inducing tilts in these illusions.

---

[6] This analytical approach can be replaced by any other application-based method to extract the relevant features necessary from the edge map for further investigations (the edge map is a DoG/LoG at multiple scales and is actually the encoded information of the contrast inside an image). Vis-CRF model can be used at the edge detection stage of any typical application that requires more in depth edge information such as in medical and military-related imaging.

We suggest that for complex Tile Illusions with inducing tilt effects and a broader range of Geometrical illusions similar to the investigated patterns which have a diverse range of tilt and brightness/contrast cues, we need some kind of fusion of multiscale (local and global) representations of the input pattern, considering the focus point and the changes of illusory effect with saccade. In parallel to this, a holistic view of the pattern for a complete explanation is required. A psychophysical assessment of the model predictions will also help to validate the results as well as towards the design of an analytical model to search for different visual clues in natural or illusion patterns similar to our visual processing.

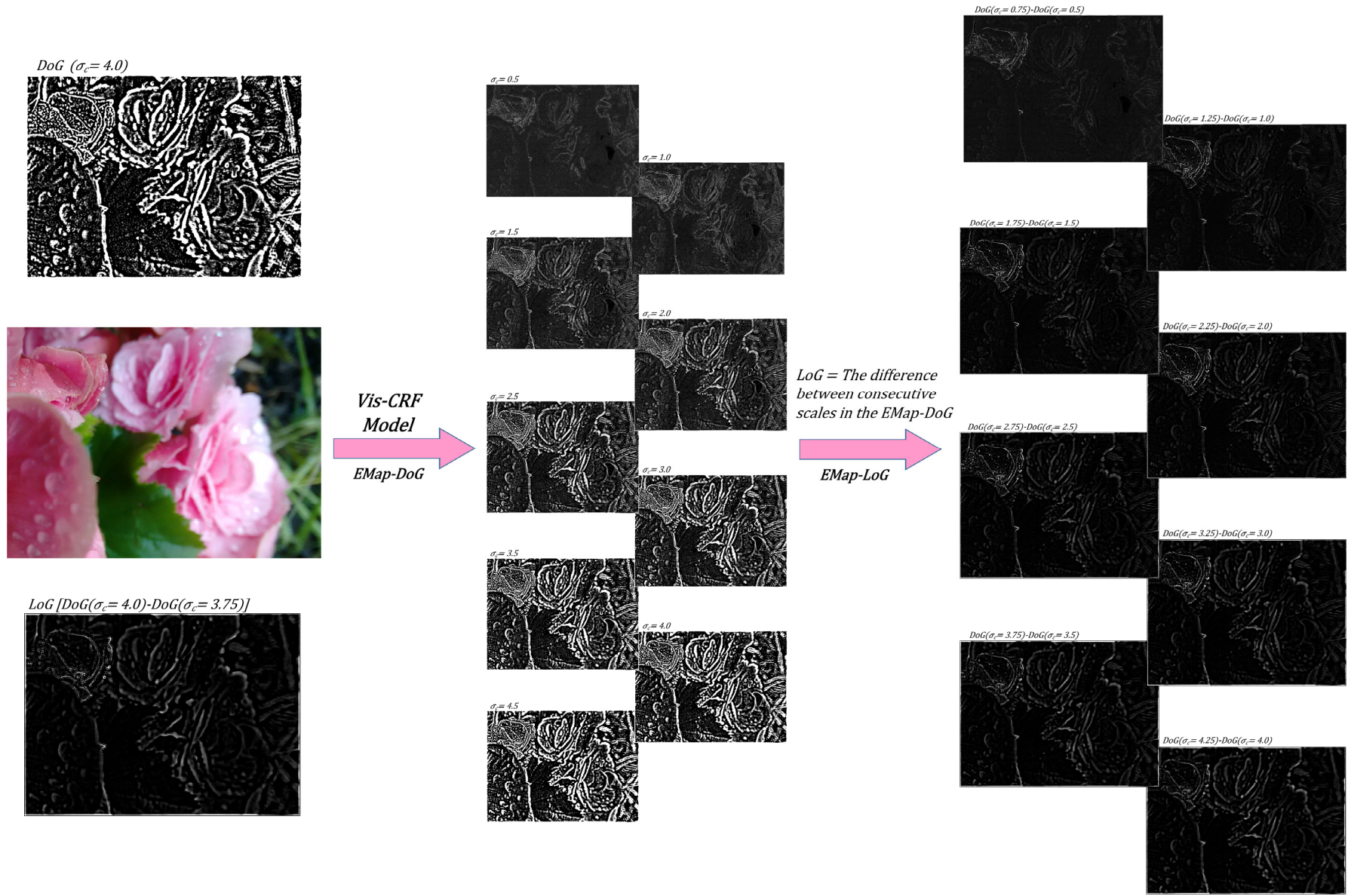

**Figure 23** Two edge map layers in our Vis-CRF model: EMap-DoG in the middle and EMap-LoG on the right. The range of scales are from $0.25$ to $10$ with incremental steps of $0.25$ for a complete analysing of the image. A limited number of scales are shown in the figure due to shortage of space ($0.5$ to $4.5$). LoG was calculated at the second layer by finding the differences between two consecutive scales in the DoG edge map, rather than using the LoG formula and will add complexity to the model. This is feasible since in the EMap-DoG, there is a gradual change in the scale of DoGs and the encoding of visual input. LoG highlighting the most dominant edges (in our On-Cell Vis-CRF model). Two enlarged versions for a particular scale of DoG ($\sigma_c = 4.0$) and one level of LoG are shown in the left next to the image.

*EMap-LoG*

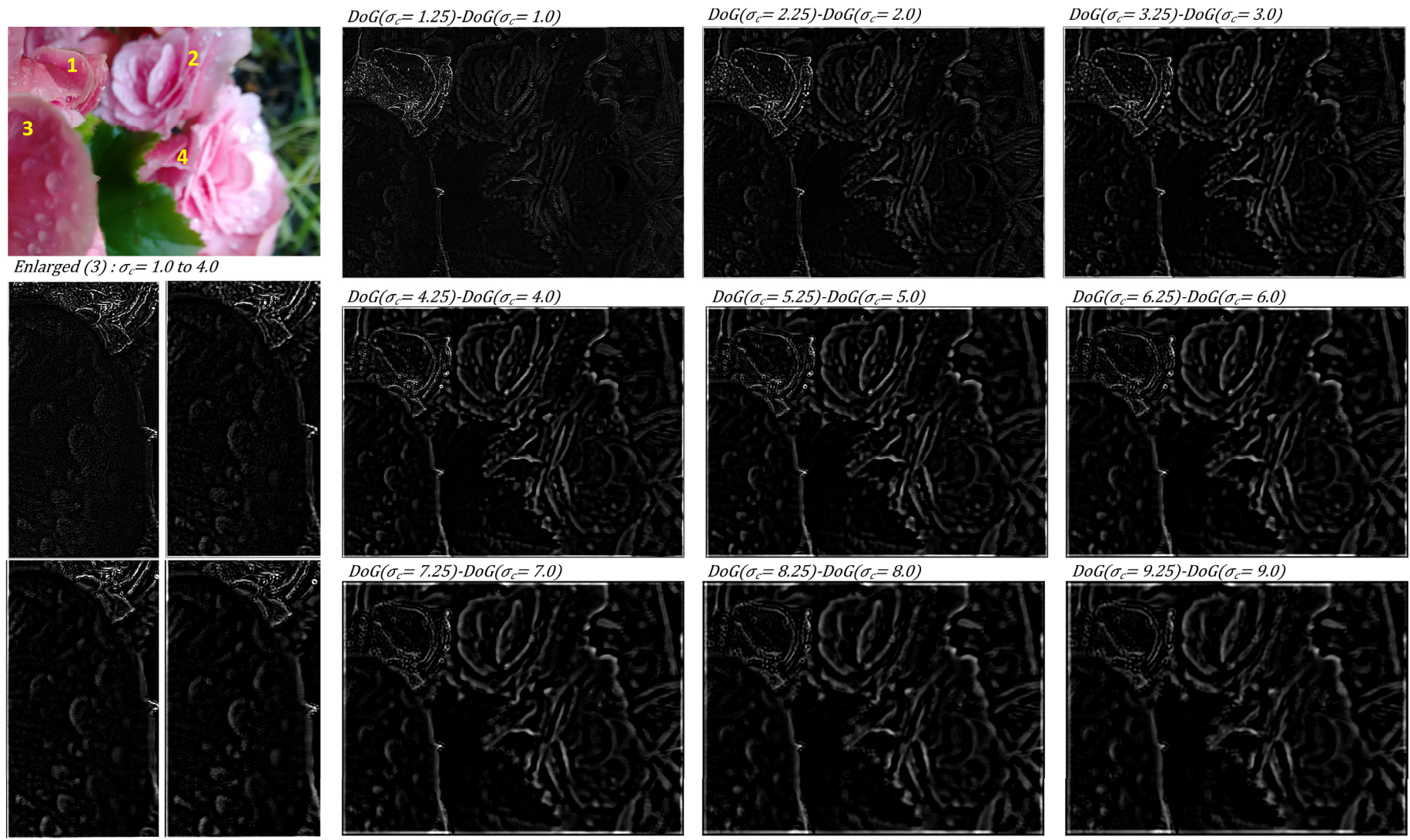

**Figure 24** LoG edge map (EMap-LoG) presented at nine different scales (the differences in between DoG scales from $\sigma_c = 1$ to $9$ with incremental steps of $1$) for the sample natural image on the top left. The bottom left corner shows the enlarged cropped sections from Rose#3 and Rose#1at the first four scales. Vis-CRF model can detect dominant edges in the pattern where the edges are clear in the LoG edge map as well as in the DoG edge map. We can see the sharp edges in Rose#1 at fine scales as well as some edges that revealed for the dew on top of Rose#3 although the image at this part is quite blurry. If an edge appears across multiple scales of the LoG edge map it is a dominant edge. The image size is $2250 \times 1644$px, and the fundamental parameters of the model are $s = 1.6$, $h = 8$ (Surround and Window ratios respectively).

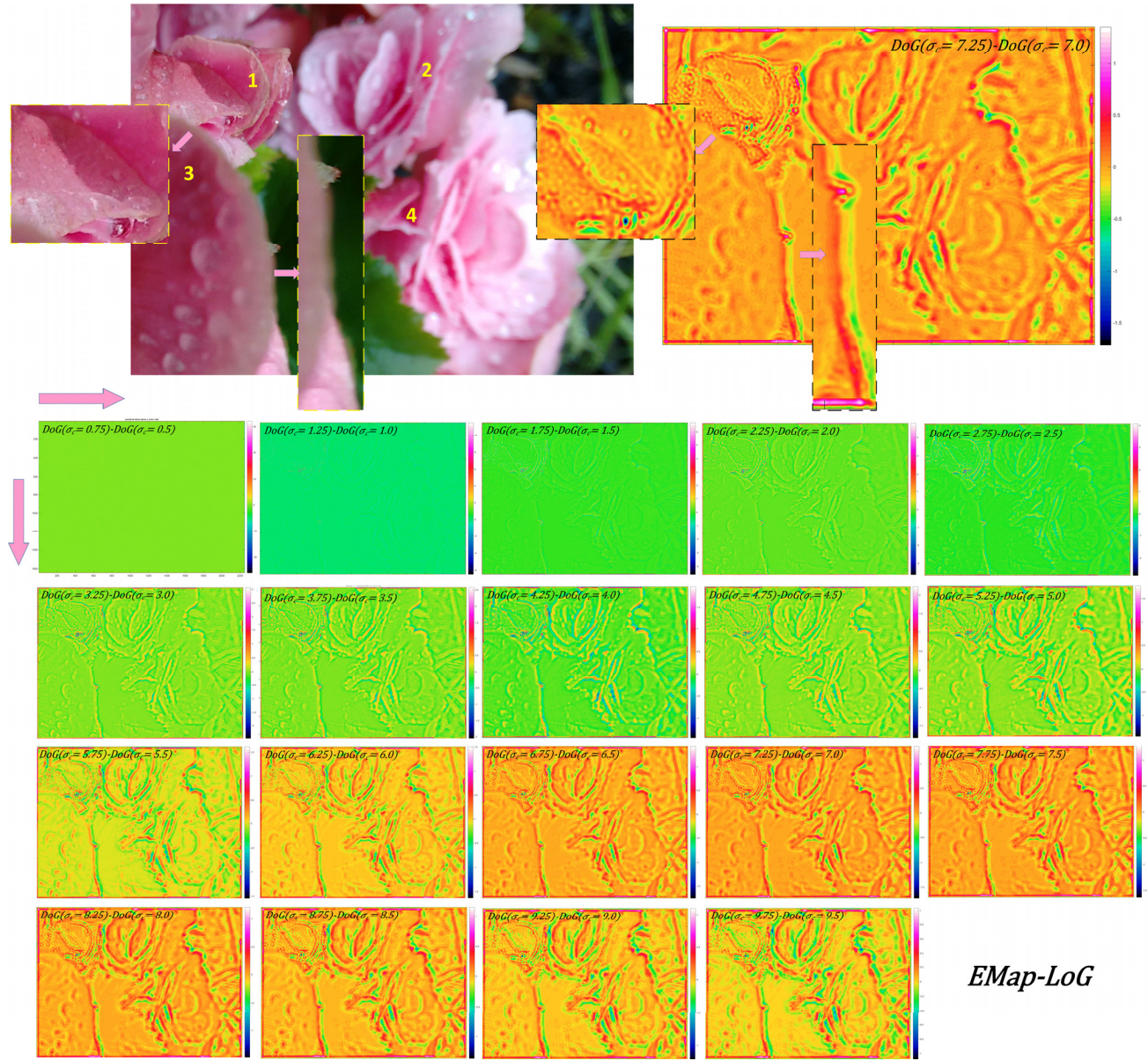

**Figure 25** LoG edge map in Jetwhite colormap (Powers, 2016) is presented at nineteen different scales here (The scale parameter is from $0.25$ to $10.0$ with incremental steps of $0.25$ in our complete analysis but a limited range are shown in the figure) for the sample natural image given on the top left. The LoG edge map is calculated by finding the differences between consecutive scales in the EMap-DoG. An enlarged version of the LoG around scale seven ($\sigma_c = 7.0$) for this pattern has shown on the top right next to the image. The enlarged versions of two cropped samples from Rose#1 and Rose #3 are provided at the top, letting us to compare the difference between a blurry edge in Rose#3 and sharp edges in Rose#1 at one level of the LoG edge map. Vis-CRF can detect the blurry edges of Rose#3 petal as well as Rose#1 petal structure at fine to medium scales. The fundamental parameters of the model are $s = 1.6$, $h = 8$ (Surround and Window ratios respectively).

# References

Barlow, HB. 1969. 'Pattern recognition and the responses of sensory neurons', *Ann N Y Acad Sci*, 156: 872-81.

Barlow, HB, Derrington, AM, Harris, LR, and Lennie, P. 1977. 'The effects of remote retinal stimulation on the responses of cat retinal ganglion cells', *J Physiol*, 269: 177-94.

Barlow, HB, and Hill, RM. 1963. 'Evidence for a Physiological Explanation of the Waterfall Phenomenon and Figural after-Effects', *Nature*, 200: 1345-7.

Basu, M. 2002. 'Gaussian-based edge-detection methods-a survey. IEEE Transactions on Systems, Man, and Cybernetics, Part C Applications and Reviews , 32(3), 252-260'.

Bekesy Von, G. 1967. 'Sensory inhibition'.

Blakeslee, B, and McCourt, ME. 1999. 'A multiscale spatial filtering account of the White effect, simultaneous brightness contrast and grating induction', *Vision Res*, 39: 4361-77.

Bressan, P. 2006. 'The place of white in a world of grays: a double-anchoring theory of lightness perception', *Psychol Rev*, 113: 526-53.

Burt, PJ, and Adelson, EH. 1983. 'The Laplacian Pyramid as a Compact Image Code', *Ieee Transactions on Communications*, 31: 532-40.

Campbell, FW, and Robson, JG. 1968. ' Application of Fourier analysis to the visibility of gratings ', *The Journal of physiology, 197(3), 551-566*.

Carandini, M. 2004a. 'Receptive Fields and Suppressive Fields in the Early Visual System', *Cognitive Neurosciences Iii, Third Edition*, 3: 313-26.

Carandini, M. 2004b. 'Receptive Fields and Suppressive Fields in the Early Visual System', *Cognitive Neurosciences Iii, Third Edition*, 313: 313-26.

Cavanaugh, JR, Bair, W, and Movshon, JA. 2002. 'Nature and interaction of signals from the receptive field center and surround in macaque V1 neurons', *J Neurophysiol*, 88: 2530-46.

Cleland, BG, Levick, WR, and Sanderson, KJ. 1973. 'Properties of sustained and transient ganglion cells in the cat retina', *J Physiol*, 228: 649-80.

Craft, E, Schutze, H, Niebur, E, and von der Heydt, R. 2007. 'A neural model of figure-ground organization', *J Neurophysiol*, 97: 4310-26.

Croner, LJ, and Kaplan, E. 1995. 'Receptive fields of P and M ganglion cells across the primate retina', *Vision Res*, 35: 7-24.

Daugman, JG. 1980. 'Two-dimensional spectral analysis of cortical receptive field profiles', *Vision Res*, 20: 847-56.

Earle, D, and Maskell, S. 1993. 'Twisted Cord and reversal of Cafe Wall Illusion', *Perception, 22(4), 383-390*.

Economou, E, Zdravkovic, S, and Gilchrist, A. 2007. 'Anchoring versus spatial filtering accounts of simultaneous lightness contrast', *J Vis*, 7: 2 1-15.

Enroth-Cugell, C, and Robson, JG. 1966. 'The contrast sensitivity of retinal ganglion cells of the cat', *J Physiol*, 187: 517-52.

Enroth-Cugell, C, and Shapley, RM. 1973. 'Adaptation and dynamics of cat retinal ganglion cells', *J Physiol*, 233: 271-309.

Field, GD, and Chichilnisky, EJ. 2007. 'Information processing in the primate retina: circuitry and coding', *Annu Rev Neurosci*, 30: 1-30.

Flicker. 'Splendid Spiral: Flicker photo share', *https://www.pinterest.com.au/pin/40229817919848034 2/visual-search/*.

Frishman, LJ, and Linsenmeier, RA. 1982. 'Effects of picrotoxin and strychnine on non-linear responses of Y-type cat retinal ganglion cells', *J Physiol*, 324: 347-63.

Gauthier, JL, Field, GD, Sher, A, Greschner, M, Shlens, J, Litke, AM, and Chichilnisky, EJ. 2009. 'Receptive fields in primate retina are coordinated to sample visual space more uniformly', *PLoS Biol*, 7: e1000063.

Ghosh, K, Sarkar, S, and Bhaumik, K. 2006. 'Early vision and image processing: evidences favouring a dynamic receptive field model', *In Computer Vision, Graphics and Image Processing. Springer Berlin Heidelberg*: 216-27.

Ghosh, K, Sarkar, S, and Bhaumik, K. 2007. 'Understanding image structure from a new multi-scale representation of higher order derivative filters', *Image and Vision Computing*, 25: 1228-38.

Gilchrist, A, Kossyfidis, C, Bonato, F, Agostini, T, Cataliotti, J, Li, X, ..., and Economou, E. 1999. 'An anchoring theory of lightness perception', *Psychological Review*, 106(4): 795.

Gollisch, T, and Meister, M. 2010. 'Eye smarter than scientists believed: neural computations in circuits of the retina', *Neuron*, 65.2 150-64.

Gregory, RL, and Heard, P. 1979. 'Border locking and the Cafe Wall illusion', *Perception*, 8: 365-80.

Hubel, DH, and Wiesel, TN. 1962. 'Receptive fields, binocular interaction and functional architecture in the cat's visual cortex', *J Physiol*, 160: 106-54.

Jacques, L, Duval, L, Chaux, C, and Peyré, G. 2011. 'A panorama on multiscale geometric representations,

intertwining spatial, directional and frequency selectivity', *Signal Processing*, 91: 2699-730.

Jameson, D. 1985. 'Opponent-colours theory in the light of physiological findings', *Central and peripheral mechanisms of colour vision*: 83-102.

Jameson, D, and Hurvich, LM. 1989. 'Essay concerning color constancy', *Annu Rev Psychol*, 40: 1-22.

Kitaoka, A. 1998. 'A Bulge', *http://www.psy.ritsumei.ac.jp/~akitaoka/Bulge02L.jpg* AND *http://www.ritsumei.ac.jp/~akitaoka/index-e.html*.

Kitaoka, A. 2000. 'Trampoline pattern ', *http://www.psy.ritsumei.ac.jp/~akitaoka/trampolineL.jpg* AND *http://www.ritsumei.ac.jp/~akitaoka/motion-e.html*.

Kitaoka, A. 2007. 'Tilt illusions after Oyama (1960): A review', *Japanese Psychological Research*, 49: 7-19.

Kuffler, SW. 1952. 'Neurons in the retina; organization, inhibition and excitation problems', *Cold Spring Harb Symp Quant Biol*, 17: 281-92.

Lindeberg, T. 1994. 'Scale-space theory: A basic tool for analyzing structures at different scales. Journal of applied statistics, 21(1-2), 225-270'.

Lindeberg, T. 2011. 'Generalized Gaussian scale-space axiomatics comprising linear scale-space, affine scale-space and spatio-temporal scale-space', *Journal of Mathematical Imaging and Vision*, 40(1): :36-81.

Lindeberg, T, and Florack, L. 1994. 'Foveal scale-space and the linear increase of receptive feld size as a function of eccentricity', *KTH Royal Institute of Technology*.

Linsenmeier, RA, Frishman, LJ, Jakiela, HG, and Enroth-Cugell, C. 1982. 'Receptive field properties of x and y cells in the cat retina derived from contrast sensitivity measurements', *Vision Res*, 22: 1173-83.

Logan Jr, BF. 1977. 'Information in the zero crossings of bandpass signals', *Bell System Technical Journal, 56(4), 487-510.*

Logvinenko, AD, and Kane, J. 2004. 'Hering's and Helmholtz's types of simultaneous lightness contrast', *J Vis*, 4: 1102-10.

Lourens, T. 1995. 'Modeling retinal high and low contrast sensitivity filters', *From Natural to Artificial Neural Computation*, 930: 61-68.

Lowe, DG. 1999. 'Object recognition from local scale-invariant features', *Computer vision and Image Understanding. The proceedings of the seventh IEEE international conference on*, 2: 1150-57 vol.2.

Lulich, DP, and Stevens, KA. 1989. 'Differential contributions of circular and elongated spatial filters to the Cafe Wall illusion', *Biol Cybern*, 61: 427-35.

Lv, YQ, Jiang, GY, Yu, M, Xu, HY, Shao, F, and Liu, SS. 2015. 'Difference of Gaussian Statistical Features Based Blind Image Quality Assessment: A Deep Learning Approach', *2015 Ieee International Conference on Image Processing (Icip)*: 2344-48.

Mallat, S. 1996. 'Wavelets for a vision', *Proceedings of the IEEE*, 84: 604-14.

Marr, D. 1976. 'Early processing of visual information', *Philos Trans R Soc Lond B Biol Sci*, 275: 483-519.

Marr, D. 1982. 'Vision. A computational investigation into the human representation and processing of visual information ', *WH San Francisco: Freeman and Company, 1(2)*.

Marr, D, and Hildreth, E. 1980. 'Theory of edge detection', *Proc R Soc Lond B Biol Sci*, 207: 187-217.

Marr, D, and Poggio, T. 1979. 'A computational theory of human stereo vision', *Proceedings of the Royal Society of London. Series B. Biological Sciences, 204(1156), 301-328.*

Marr, D, and Ullman, S. 1981. 'Directional Selectivity and Its Use in Early Visual Processing', *Proceedings of the Royal Society Series B-Biological Sciences*, 211: 151-+.

Marr, D, Ullman, S, and Poggio, T. 1979. 'Bandpass channels, zero-crossings, and early visual information processing', *J Opt Soc Am*, 69: 914-6.

Martinez-Conde, S, Macknik, SL, and Hubel, DH. 2004. 'The role of fixational eye movements in visual perception', *Nat Rev Neurosci*, 5: 229-40.

Nematzadeh, N. 2018. 'A Neurophysiology Model that Makes Quantifiable Predictions of Geometric Visual Illusions', *Ph.D. Dissertation. Flinders University of South Australia*.

Nematzadeh, N, Lewis, TW, and Powers, DMW. 2015. 'Bioplausible multiscale filtering in retinal to cortical processing as a model of computer vision', *ICAART2015-International Conference on Agents and Artificial Intelligence. SCITEPRESS*.

Nematzadeh, N, and Powers, DM. 2017a. 'The Cafe Wall Illusion: Local and Global Perception from multiple scale to multiscale', *Applied Computational Intelligence and Soft Computing, Special issue of Imaging, Vision, and Pattern Recognition*.

Nematzadeh, N, Powers, DM, and Lewis, TW. 2019. 'Informing Computer Vision with Optical Illusions. ', *arXiv preprint arXiv:1902.02922. Submitted to Brain Informatics journal*.

Nematzadeh, N, and Powers, DMW. 2016a. 'A bioplausible model for explaining Café Wall illusion: foveal vs peripheral resolution', *In International Symposium on Visual Computing. Springer International Publishing*: 426-38.

Nematzadeh, N, and Powers, DMW. 2016b. 'A quantitative analysis of tilt in the Cafe Wall illusion: a bioplausible model for foveal and peripheral vision', *2016 International Conference on Digital Image Computing: Techniques and Applications (Dicta)*: 23-30.

Nematzadeh, N, and Powers, DMW. 2017b. 'The Cafe Wall Illusion: Local and Global Perception from multiple scale to multiscale', *Journal of Applied Computational Intelligence and Soft Computing: Special issue of Imaging, Vision, and Pattern Recognition*.

Nematzadeh, N, and Powers, DMW. 2017c. 'A Predictive Account of Café Wall Illusions Using a Quantitative Model', *submitted; arXiv preprint:1705.06846*.

Nematzadeh, N, and Powers, DMW. 2019. 'Quantified Measurement of the Tilt Effect in a Family of Café Wall Illusions', *MODVIS2019 - Session 1: Objects and Contours, https://docs.lib.purdue.edu/modvis/2019/session01/4/*.

Nematzadeh, N, Powers, DMW, and Lewis, TW. 2016. 'Quantitative Analysis of a Bioplausible Model of Misperception of Slope in the Café Wall Illusion', *In Asian Conference on Computer Vision. Springer, Cham*: 622-37.

Nematzadeh, N, Powers, DMW, and Lewis, TW. 2017. 'Bioplausible multiscale filtering in retino-cortical processing as a mechanism in perceptual grouping', *Brain informatics, 4(4), 271-293*.

Powers, DMW. 1983. 'Lateral interaction behaviour derived from neural packing considerations', *University of New South Wales, School of Electrical Engineering and Computer Science, Sydney*.

Powers, DMW. 2016. 'Jetwhite color map. Mathworks – https://au.mathworks.com/matlabcentral/fileexchange/48419-jetwhite-colours-/content/jetwhite.m. '.

Ratliff, F, Knight, BW, and Graham, N. 1969. 'On Tuning and Amplification by Lateral Inhibition', *Proceedings of the National Academy of Sciences*, 62: 733-40.

Robson, JG. 1983. 'Frequency domain visual processing', *Physical and biological processing of images, Springer Berlin Heidelberg*: 73-87.

Rodieck, RW, and Stone, J. 1965. 'Analysis of Receptive Fields of Cat Retinal Ganlion Cells', *Journal of Neurophysiology, 28(5), 833-849*.

Romeny, BM. 2008. 'Front-end vision and multi-scale image analysis: multi-scale computer vision theory and applications, written in mathematica (Vol. 27). Springer Science & Business Media.'.

Romeny, BMH. 2003. 'Front-end vision and multi-scale image analysis: multi-scale computer vision theory and applications, written in Mathematica'.

Rosenfeld, A, and Thurston, M. 1971. 'Edge and Curve Detection for Visual Scene Analysis', *IEEE Transactions on computers, 100(5), 562-569*.

Roska, B, and Werblin, F. 2003. 'Rapid global shifts in natural scenes block spiking in specific ganglion cell types', *Nat Neurosci*, 6: 600-8.

Schiller, PH. 2010. 'Parallel information processing channels created in the retina', *Proc Natl Acad Sci U S A*, 107: 17087-94.

Shapley, R, and Perry, VH. 1986. 'Cat and Monkey Retinal Ganglion-Cells and Their Visual Functional Roles', *Trends in Neurosciences*, 9: 229-35.

Smith, SW. 2003. 'Digital signal processing: a practical guide for engineers and scientists', *Elsevier*.

So, YT, and Shapley, R. 1981. 'Spatial tuning of cells in and around lateral geniculate nucleus of the cat: X and Y relay cells and perigeniculate interneurons', *J Neurophysiol*, 45: 107-20.

Taubman, D, and Marcellin, M. 2012. 'JPEG2000 image compression fundamentals, standards and practice: image compression fundamentals, standards and practice', *(Vol. 642). Springer Science & Business Media*.

Todorovic, DM. 2017. 'A Computational Account of a Class of Orientation Illusions', *MODVIS2017 - Computational and Mathematical Models in Vision*.

von der Malsburg, C. 1973. 'Self-organization of orientation sensitive cells in the striate cortex', *Kybernetik*, 14: 85-100.

Wei, H, Zuo, Q, and Lang, B. 2011. 'Multi-scale Image Analysis Based on Non-Classical Receptive Field Mechanism', *Springer Berlin Heidelberg*, 7064: 601-10.

Weng, S, Sun, W, and He, S. 2005. 'Identification of ON-OFF direction-selective ganglion cells in the mouse retina', *J Physiol*, 562: 915-23.

Wilson, HR, and Giese, SC. 1977. 'Threshold visibility of frequency gradient patterns', *Vision Research, 17(10), 1177-1190*.

Young, RA. 1985. 'The Gaussian derivative theory of spatial vision: Analysis of cortical cell receptive field line-weighting profiles': 48090-9055.

Young, RA. 1987. 'The Gaussian derivative model for spatial vision: I. Retinal mechanisms', *Spat Vis*, 2: 273-93.